\RequirePackage[svgnames]{xcolor}

\documentclass[11pt,letterpaper]{mystyle}

\usepackage[all]{hypcap}
\usepackage[svgnames]{xcolor}
\usepackage[comma,authoryear,compress]{natbib}
\usepackage{hyperref}[citecolor=lightblue]

\hypersetup{
    colorlinks = true,
    citecolor = {YaleBlue},
}

\usepackage{algorithm}
\usepackage{algorithmicx}
\usepackage{algpseudocode}
\usepackage{microtype}
\usepackage{graphicx}
\expandafter\def\csname ver@subfig.sty\endcsname{}
\usepackage{booktabs} %
\usepackage{float}
\usepackage{bigstrut}

\usepackage{amsmath}
\usepackage{amssymb}
\usepackage{mathtools}
\usepackage{amsthm}
\usepackage{mathrsfs}
\usepackage{nicefrac}
\usepackage{dsfont}
\usepackage{enumitem}
\usepackage{subcaption}
\usepackage{graphicx,subfig}
\usepackage{cleveref}
\usepackage{bxcoloremoji}

\usepackage{float}

\usepackage[utf8]{inputenc} %
\usepackage[T1]{fontenc}    %
\usepackage{hyperref}       %
\usepackage{url}            %
\usepackage{booktabs}       %
\usepackage{amsfonts}       %
\usepackage{nicefrac}       %
\usepackage{microtype}      %
\usepackage{graphicx}
\usepackage{subcaption} %
\usepackage{amssymb}
\usepackage{fdsymbol}
\usepackage{wrapfig}
\usepackage{lipsum}
\usepackage{enumitem}
\usepackage{stackengine}
\usepackage[font=small,labelfont=bf]{caption}
\usepackage{color}
\usepackage{adjustbox}

\usepackage{rotating}
\usepackage{makecell}

\usepackage{multirow}

\newcommand{\x}{\mathbf{x}}

\definecolor{blanchedalmond}{rgb}{1.0, 0.92, 0.8}
\definecolor{carmine}{rgb}{0.59, 0.0, 0.09}
\definecolor{lightblue}{rgb}{0.22,0.45,0.70}%

\renewcommand{\mathbf}{\boldsymbol}

\makeatletter
\def\Ddots{\mathinner{\mkern1mu\raise\p@
\vbox{\kern7\p@\hbox{.}}\mkern2mu
\raise4\p@\hbox{.}\mkern2mu\raise7\p@\hbox{.}\mkern1mu}}
\makeatother

\definecolor{amaranth}{rgb}{0.9, 0.17, 0.31}
\definecolor{antiquebrass}{rgb}{0.8, 0.58, 0.46}
\definecolor{antiquefuchsia}{rgb}{0.57, 0.36, 0.51}
\definecolor{chromeyellow}{rgb}{0.31, 0.47, 0.26}

\newcommand{\github}{\raisebox{-1.5pt}{\includegraphics[height=1.05em]{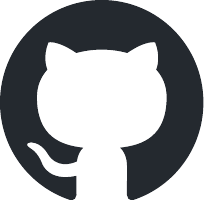}}}

\newtcolorbox{AIbox}[2][]{aibox,title=#2,#1}
\definecolor{lightblue}{rgb}{0.22,0.45,0.70}%
\definecolor{Gray}{gray}{0.95}
\definecolor{Cornsilk}{rgb}{1.0, 0.97, 0.86}

\usepackage{amsmath}

\usepackage[all]{hypcap}

\title{MuRF: Unlocking the Multi-Scale Potential of Vision Foundation Models}

\runningtitle{MuRF: Unlocking the Multi-Scale Potential of Vision Foundation Models}

\author{
  Bocheng Zou$^{\dagger}$,
  Mu Cai$^{\dagger}$,
  Mark Stanley$^{*}$,
  Dingfu Lu$^{*}$,
  and Yong Jae Lee
}

\affil[1]{University of Wisconsin-Madison}

\correspondingauthor{
$^{\dagger}$: Equal Contribution, $^{*}$: Equal Second Author}

\newcommand{\cmark}{\ding{51}}%
\newcommand{\xmark}{\ding{55}}%

\newcommand{\methodsfullname}{\textbf{Mu}lti-\textbf{R}esolution \textbf{F}usion}
\newcommand{\methodshortname}{\textit{MuRF}}
\newcommand{\shortname}{\textit{MuRF}}

\begin{document}

\begin{abstract}

Vision Foundation Models (VFMs) have become the cornerstone of modern computer vision, offering robust representations across a wide array of tasks. While recent advances allow these models to handle varying input sizes during training, inference typically remains restricted to a single, fixed scale. This prevalent single-scale paradigm overlooks a fundamental property of visual perception: varying resolutions offer complementary inductive biases, where low-resolution views excel at global semantic recognition and high-resolution views are essential for fine-grained refinement. In this work, we propose \methodsfullname{} (\methodshortname{}), a simple yet universally effective strategy to harness this synergy at inference time. Instead of relying on a single view, \methodshortname{} constructs a unified representation by processing an image at multiple resolutions through a frozen VFM and fusing the resulting features. 
The universality of \shortname{} is its most compelling attribute. It is not tied to a specific architecture, serving instead as a fundamental, training-free enhancement to visual representation. We empirically validate this by applying \shortname{} to a broad spectrum of critical computer vision tasks across multiple distinct VFM families—primarily DINOv2, but also demonstrating successful generalization to contrastive models like SigLIP2. 

\vspace{2em}

\coloremojicode{1F4C5} \textbf{Date}: March 26, 2026

\coloremojicode{1F3E0} \textbf{Projects}: \href{https://MuRF-VFM.github.io}{https://MuRF-VFM.github.io}

\github{} \textbf{Code Repository}: \href{https://github.com/orgs/MuRF-VFM/repositories}{https://github.com/orgs/MuRF-VFM}

\coloremojicode{1F4E7} \textbf{Contact}: \href{mailto:bochengz@cs.wisc.edu}{bochengz@cs.wisc.edu}

\end{abstract}

\maketitle
\vspace{3mm}

\section{Introduction}
\label{sec:intro}
The landscape of computer vision has been reshaped by standardizing visual representation learning through large-scale Vision Foundation Models (VFMs)~\citep{oquab2024dinov}. These models, pre-trained on vast datasets, provide widely transferable features that serve as the bedrock for numerous downstream applications. A key axis of evolution in this domain has been handling image resolution. While early ViT-based approaches required rigid, fixed-size inputs~\citep{dosovitskiy2021an}, recent paradigms like ``multi resolution'' training (e.g., DINOv2~\citep{oquab2024dinov}%
) have enabled models to process images of arbitrary aspect ratios and scales flexibly.

\begin{figure}[t]
    \centering
    \begin{subfigure}[t]{0.49\linewidth}
        \centering
        \includegraphics[width=\linewidth]{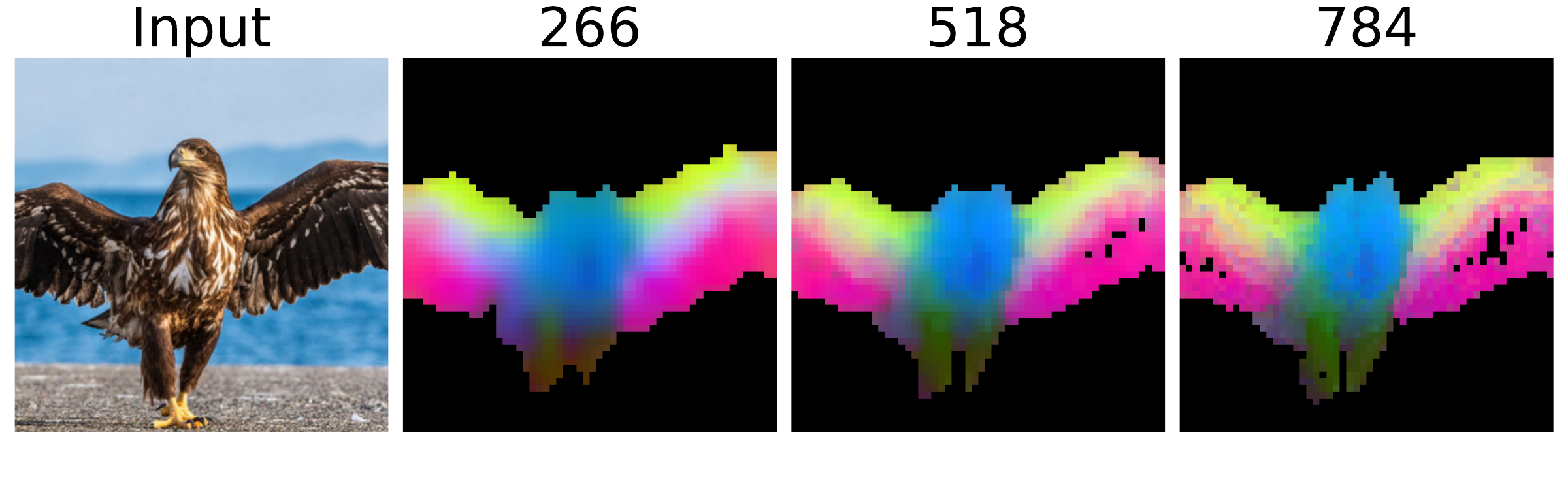}
    \end{subfigure}
    \hfill
    \begin{subfigure}[t]{0.49\linewidth}
        \centering
        \includegraphics[width=\linewidth]{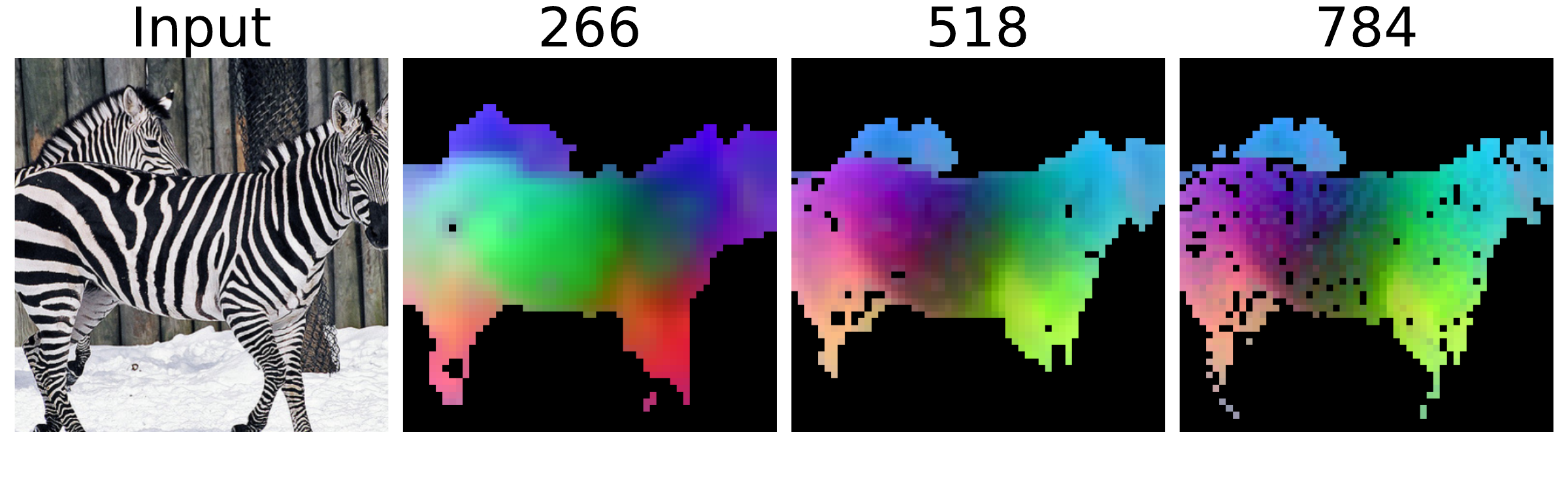}
    \end{subfigure}

    \caption{The ``Recognition vs. Refinement'' Dynamic. The feature map obtained when the input is resized to 266, 518, 784. At lower resolutions, the representation is globally coherent, enabling robust \textbf{recognition}. At higher resolutions, boundary details are sharper, enabling precise \textbf{refinement}, but the object's interior becomes noisy, risking incomplete segmentation. Our work is motivated by synergizing these two roles.}
    \label{fig:intro_pca}
\end{figure}

Despite this flexibility in \textit{training}, standard \textit{inference} protocols remain surprisingly rigid. Typically, an input image is resized to a single ``optimal'' resolution before being processed by the VFM. We argue that this single-scale inference discards a wealth of information inherent in the multi-scale nature of visual data. As shown in Figure~\ref{fig:intro_pca}, it is well-understood that a ``division of labor'' exists across different viewing scales: low-resolution inputs, by virtue of their larger relative patch sizes, excel at capturing globally coherent semantic context (recognition). High-resolution inputs are indispensable for resolving fine-grained high-frequency details and precise boundaries (refinement). By restricting inference to any single scale, current methods inevitably compromise either global coherence or local precision.

We posit that unlocking the full potential of pre-trained VFMs requires actively aggregating these complementary views at inference time. To this end, we introduce \methodsfullname{} (\methodshortname{}), a general and straightforward strategy designed to create a scale-robust visual representation. Rather than modifying the VFM backbone, \methodshortname{} treats it as a frozen feature extractor. It processes the same input image across a pyramid of resolutions and fuses the resulting features into a unified representation. This simple operation effectively synergizes the recognition strengths of low-resolution views with the refinement capabilities of high-resolution views, without requiring expensive multi-scale training of the backbone itself.

The universality of \shortname{} is its most compelling attribute. It is not tied to a specific architecture or task but serves as a fundamental enhancement to the visual representation. We empirically validate this by applying \shortname{} to a broad spectrum of critical computer vision tasks, all using frozen VFM backbones:
\begin{itemize}
    \item \textbf{Dense Prediction (Probing):} In semantic segmentation and depth estimation, simple linear heads trained on \shortname{} representations significantly outperform those trained on standard single-scale features, proving that multi-scale fusion enriches basic feature quality.
    \item \textbf{Multimodal Understanding:} When used as the visual encoder for Multimodal Large Language Models (MLLMs) in tasks like Visual Question Answering (VQA), \shortname{} provides LLMs with a more comprehensive visual context, enabling them to reason about both macroscopic scenes and microscopic details simultaneously.
    \item \textbf{Unsupervised Anomaly Detection:} In this training-free regime, \shortname{} effectively resolves the trade-off between detecting large structural anomalies and tiny surface defects, achieving state-of-the-art performance on demanding benchmarks like MVTec AD 2 TEST\textsubscript{priv,mix}~\citep{heckler2025mvtec}.
\end{itemize}
Our results suggest that effectively leveraging multi-resolution views at inference time is a general principle that consistently improves VFM-based systems. \shortname{} offers a simple, unified, and highly effective implementation of this principle.

\begin{figure*}[t]
    \centering
    \includegraphics[width=\textwidth]{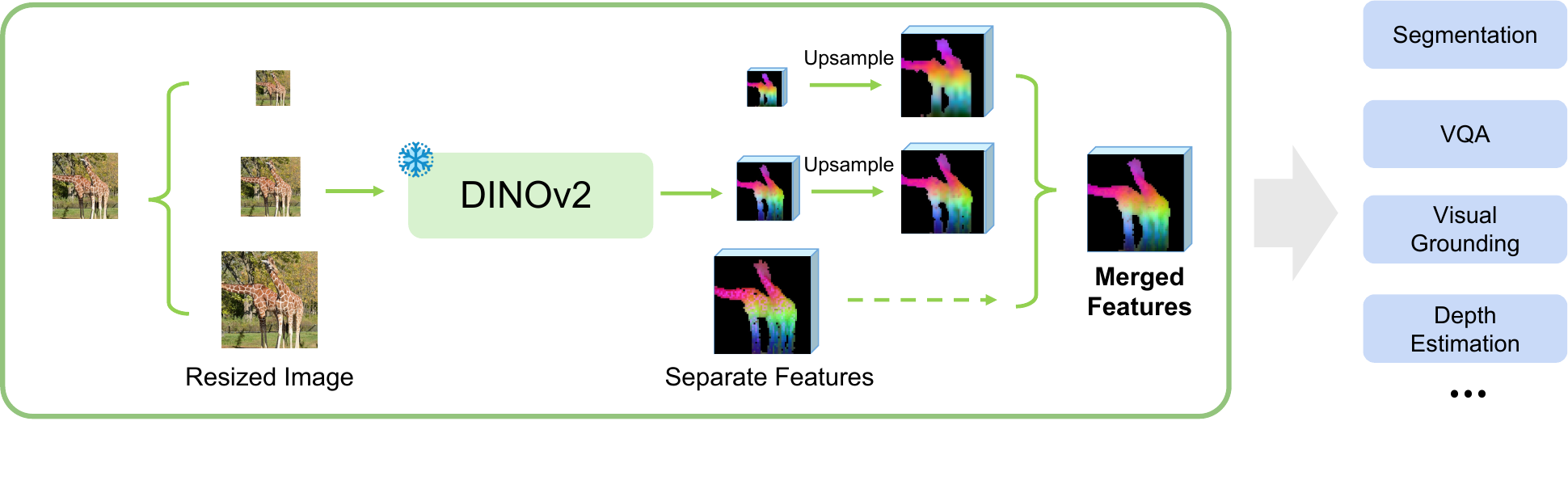}
    \caption{Overview of \methodsfullname{} (\shortname{}). An input image is resized to multiple resolutions and each view is processed by a frozen DINOv2 encoder to produce separate feature maps. These features are upsampled to a shared spatial resolution and fused into a single multi-resolution representation, which can then be used by lightweight task-specific heads for semantic segmentation, depth estimation, visual question answering, visual grounding, and other downstream tasks. Background is removed from the PCA figures.}
    \label{fig:teaser}
\end{figure*}

\section{Related Work}

Our work explores unlocking the full potential of frozen VFMs by revisiting and modernizing multi-resolution inference strategies. We review the evolution of VFMs regarding standard input resolutions and discuss historical and contemporary approaches to multi-scale representation learning.

\subsection{Vision Foundation Models and Input Resolution}
The field of computer vision has been fundamentally altered by the rise of Vision Foundation Models (VFMs). Initially dominated by Convolutional Neural Networks (CNNs) like ResNet~\citep{He_2016_CVPR} trained via supervised learning, the paradigm shifted toward Vision Transformers (ViTs)~\citep{dosovitskiy2021an} leveraging large-scale self-supervision. Models such as CLIP~\citep{pmlr-v139-radford21a} and DINO/DINOv2~\citep{Caron_2021_ICCV, oquab2024dinov} now serve as ubiquitous feature extractors across diverse tasks.

Traditionally, ViTs have been constrained by rigid input resolutions (e.g., $224 \times 224$) due to their reliance on fixed positional embeddings. This limitation often necessitates aggressive resizing or cropping, potentially discarding critical visual information. Recent advancements, such as DINOv2~\citep{oquab2024dinov}, NaViT~\citep{NEURIPS2023_06ea400b} and FlexViT~\citep{Beyer_2023_CVPR}, have introduced multi resolution or native resolution training, allowing models to process images in %
varying sizes during training.

While multi resolution \textit{training} enhances model flexibility, standard \textit{inference} protocols typically revert to utilizing a single, fixed scale, often the highest resolution practically affordable. This single-scale inference overlooks the inherent ``division of labor" in visual perception: low-resolution views excel at global semantic recognition, while high-resolution views are necessary for fine-grained refinement. Our \methodsfullname{} (\methodshortname{}) strategy addresses this by explicitly aggregating features from multiple resolutions at inference time, ensuring that both global context and local details are preserved without retraining the backbone.

\subsection{Multi-Scale Representations}
Handling objects of varying scales is a classical challenge in computer vision. Early approaches relied on \textit{image pyramids}~\citep{adelson1984pyramid}, where an image is repeatedly resized and processed. While effective, this was computationally prohibitive for heavy older models. The deep learning era introduced \textit{feature pyramids}, most notably Feature Pyramid Networks (FPN)~\citep{lin2017feature}, which efficiently construct multi-scale representations within the network's forward pass.

In the era of frozen VFMs and Multimodal Large Language Models (MLLMs), explicitly constructing FPNs often requires costly task-specific training. Consequently, recent trends have shifted back towards input-level manipulations. For handling high-resolution images in MLLMs (e.g., GPT-4V~\citep{GPT4V_System_Card}, LLaVA-NeXT~\citep{liu2024llavanext}), a common technique is dividing the image into smaller, fixed-resolution tiles (patches) processed independently, often supplemented by a single low-resolution global view. Similar tiling strategies have been applied in token efficiency for multimodal LLM, such as $S^2$~\citep{10.1007/978-3-031-73242-3_25}.

While tiling allows processing arbitrarily large images, it artificially breaks continuity, making it difficult for models to reason about objects that span splitting boundaries. Furthermore, simple upsampling techniques for VFM features, like FeatUp~\citep{fu2024featup} and JAFAR~\citep{couairon2025jafar}, may recover high-frequency details but do not inherently add new information absent from the original single-scale forward pass. What's more, all of them require certain amount of training, which may lead to generalizability issues. \shortname{} revisits the classical image pyramid but modernizes it for the VFM era. By fusing multi-resolution views in the feature space, we avoid the boundary artifacts of tiling while capturing a richer, more holistic representation than single-scale upsampling, all in a completely training-free manner for the backbone.

\section{Method}

Our goal is to develop a universal, scale-robust visual representation from a frozen Vision Foundation Model (VFM) that can benefit a wide array of downstream tasks. To this end, we propose \methodsfullname{} (\methodshortname{}), a simple yet highly effective strategy for feature extraction at inference time. As illustrated in Figure~\ref{fig:teaser}, our method first constructs a rich, multi-scale feature map and then adapts this representation to specific tasks using lightweight, trainable heads.

\subsection{Multi-Resolution Feature Fusion}
\label{sec:mrf_representation}

The core of our approach is motivated by the observation that different input resolutions provide complementary visual information: low resolutions capture global context for robust \textbf{recognition}, while high resolutions provide fine-grained detail for precise \textbf{refinement}. \shortname{} explicitly harnesses this synergy by building a feature pyramid from the input space.

Given an input image $\mathbf{x} \in \mathbb{R}^{H\times W\times C}$, we first create an input pyramid by resizing it to a set of different scaling factors, $S_{\text{res}} = \{s_1, s_2, \dots, s_k\}$. This yields a collection of images $\{\mathbf{x}_s\}_{s \in S_{\text{res}}}$.

Each resized image $\mathbf{x}_s$ is then passed through a frozen VFM encoder, which we denote as $\Phi$. We extract the feature map from the encoder (typically the last layer or a specific target block). For each resolution $s$, we obtain a patch-level feature map:
\begin{equation}
    \mathcal{F}_{s} = \Phi(\mathbf{x}_s) \in \mathbb{R}^{H_{s} \times W_{s} \times d}
\end{equation}
where $(H_{s}, W_{s})$ are the spatial dimensions of the feature map for scale $s$, and $d$ is the feature (channel) dimension.

To create a single, unified representation, we fuse these individual feature maps. Each map $\mathcal{F}_{s}$ is first upsampled to a common target spatial resolution $(H', W')$, typically the original input size, using bilinear interpolation. These spatially-aligned feature maps are then concatenated along the channel dimension. This process yields the final \textbf{\methodshortname{} representation}, $\mathcal{F}_{\text{\shortname}}$:
\begin{equation}
    \mathcal{F}_{\text{\shortname}} = \text{Concat}_{s \in S_{\text{res}}} \left( \text{Upsample}(\mathcal{F}_{s}) \right) \in \mathbb{R}^{H' \times W' \times D}
    \label{eq:mrf_representation}
\end{equation}
The total channel dimension $D$ is the sum of the dimensions of the feature maps across all scales, $D = |S_{\text{res}}| \times d$. This final tensor, $\mathcal{F}_{\text{\shortname}}$, serves as a powerful, frozen representation that is spatially rich, semantically deep, and robust to scale variations.

\paragraph{Why Channel-wise Concatenation?} 
While one might consider alternative feature fusion operations—such as element-wise addition, mean pooling, or learnable attention—we intentionally select channel-wise concatenation. ViT features are highly localized and scale-dependent semantic tokens. Summation or mean pooling risks destructive interference, blending orthogonal scale-specific activations (e.g., a macroscopic semantic feature with a microscopic edge feature) into an ambiguous representation. By concatenating, we project the features into a higher-dimensional space ($D = |S_{\text{res}}| \times d$), preserving the strict independence of the ``recognition'' and ``refinement'' signals. This allows the lightweight downstream head to adaptively select and route the appropriate scale information without requiring a heavy, parameterized fusion network.

\subsection{Task-Specific Adaptation}
\label{sec:task_adaptation}

The $\mathcal{F}_{\text{\shortname}}$ representation is task-agnostic. We adapt it to various downstream tasks by attaching lightweight, task-specific heads. This allows the frozen VFM backbone to be leveraged efficiently across different domains.

\paragraph{Dense Prediction Tasks.} For tasks like semantic segmentation and depth estimation, which require a pixel-wise prediction, we attach a simple dense prediction head, $\text{Head}_{\text{dense}}(\cdot)$. This head typically consists of one or more convolutional layers (e.g., a $1\times1$ convolution) that project the $D$-dimensional feature channels to the desired number of output channels (e.g., number of semantic classes or a single depth value). The final prediction map $\hat{\mathbf{Y}} \in \mathbb{R}^{H \times W \times C_{\text{out}}}$ is obtained by:
\begin{equation}
    \hat{\mathbf{Y}} = \text{Upsample} \left( \text{Head}_{\text{dense}}(\mathcal{F}_{\text{\shortname}}) \right)
\end{equation}
During training, only the parameters of $\text{Head}_{\text{dense}}$ are updated, making the process highly efficient.

\paragraph{Unsupervised Anomaly Detection.} This task follows a training-free paradigm. Instead of fusing features into a single tensor, we extend the nearest-neighbor approach. %
For each resolution $s$, we construct a dedicated memory bank $\mathcal{M}_{s}$ from anomaly-free training images. At inference time, we compute a separate anomaly score map $\hat{S}_{s}$ for each feature map $\mathcal{F}_{s}$ by calculating the $L_2$ distance of each feature vector to its nearest neighbor in $\mathcal{M}_{s}$. The final, robust anomaly score map $\hat{\mathbf{S}}$ is produced by averaging all individual score maps after upsampling them to the original image dimensions:
\begin{equation}
    \hat{\mathbf{S}} = \frac{1}{|S_{\text{res}}|} \sum_{s \in S_{\text{res}}} \text{Upsample}(\hat{S}_{s})
\end{equation}
This fusion of scores leverages the strengths of all views for both recognition and refinement.

\paragraph{Multimodal Language Models.} In the context of MLLMs for tasks like Visual Question Answering (VQA) and visual grounding, the $\mathcal{F}_{\text{DINOv2}}$ representation serves as the visual input to the language model. The rich spatial feature map $\mathcal{F}_{\text{\shortname}}$ is first passed through a perception module or projection layer, $\text{Head}_{\text{MLLM}}(\cdot)$, which maps the visual features into the word embedding space of the LLM.
\begin{equation}
    \mathbf{E}_{\text{visual}} = \text{Head}_{\text{MLLM}}(\mathcal{F}_{\text{\shortname}})
\end{equation}
These visual embeddings $\mathbf{E}_{\text{visual}}$ are then treated as a sequence of ``visual tokens'' and prepended to the text token embeddings. By providing the LLM with a representation that is rich in both global and local detail, we empower it to answer questions that require reasoning across multiple scales. Only the MLLM's projection layer and other designated parameters are trained, while the \shortname{} feature extractor remains frozen.

\section{Experiments}

To validate the universality and effectiveness of our proposed \methodsfullname{} (\methodshortname{}) representation, we conduct a comprehensive set of experiments across four diverse and fundamental computer vision tasks: semantic segmentation, depth estimation, visual question answering, and anomaly detection. Our evaluation aims to demonstrate that \shortname{} consistently improves performance over strong, single-scale baselines, regardless of the downstream application.

\subsection{Experimental Setup}

\paragraph{VFM Backbone and \shortname{} Configuration.}

Unless otherwise specified, our \shortname{} configuration is as follows:

\begin{itemize}
    \item \textbf{Backbone:} Across all experiments, we use the publicly available, pre-trained \textbf{DINOv2-ViT-B/14}~\citep{oquab2024dinov} as our frozen VFM encoder. This ensures a fair comparison and highlights the gains attributable solely to our multi-resolution fusion strategy. We additionally experiment on \textbf{SigLIP2-Base} (Patch16 NaFlex version) ~\citep{tschannen2025siglip} to verify the applicability of our proposed approach.
    \item \textbf{Resolutions ($S_{\text{res}}$):} We are using 5 resolutions for anomaly detection, and 3 resolutions for the rest of tasks including segmentation, depth estimation, and PCA, and 2 resolutions for MLLM. We adhere to the resolution of the original implementation including LLaVA-1.5~\citep{liu2024improved} and DINOv2~\citep{oquab2024dinov}.
    
\end{itemize}
The final \shortname{} representation is generated by upsampling all extracted feature maps to a common resolution and concatenating them channel-wise, as described in Section~\ref{sec:mrf_representation}.

\paragraph{Training Details.}
For all downstream tasks (segmentation, depth estimation, and VQA), we exclusively train the task-specific heads while keeping both the VFM backbone and the \shortname{} fusion module frozen. We adhere to the optimization configurations—including optimizer, learning rate, and schedule—established in the respective original papers~\citep{oquab2024dinov, liu2024llavanext}. Additional details of the experiment for reproducibility can be found in the Supplementary Material. 

\subsection{Semantic Segmentation}

\paragraph{Setup.}
We evaluate on the challenging \textbf{ADE20k}~\citep{zhou2019semantic} and \textbf{PASCAL VOC} \citep{10.1007/s11263-009-0275-4} benchmark using the standard mean Intersection over Union (mIoU) metric. Our baseline for comparison is a linear probing setup using the same frozen DINOv2 encoder but with features extracted from a single input resolution, which represents the standard inference paradigm. All experiments are performed using DINOv2-ViT-B/14.

\begin{table*}[ht]
\centering
\caption{Combined downstream evaluation. \textbf{Semantic segmentation:} performance in mIoU (\%) on ADE20K and PASCAL VOC (higher is better). \textbf{Depth estimation:} performance on NYU Depth V2 and SUN RGB-D reporting RMSE (lower is better). \shortname{} significantly outperforms single-scale baselines, highlighting the benefit of our multi-scale representation for dense prediction and geometric reasoning. \textbf{Bold} indicates the best performance, and \underline{underline} indicates the second best. "Rel. Improv" measures the relative improvement over original DINOv2 (high resolution for semantic segmentation, and medium resolution for depth estimation)}
\label{tab:segmentation_depth_combined}
\resizebox{\textwidth}{!}{
\begin{tabular}{llcc|cccc}
\toprule
\multicolumn{2}{c}{} &
\multicolumn{2}{c}{\textbf{Semantic Segmentation $\uparrow$}} &
\multicolumn{4}{c}{\textbf{Depth Estimation $\downarrow$}} \\
\cmidrule(lr){3-4}\cmidrule(lr){5-8}
\textbf{Method} & \textbf{Arch.} &
\textbf{ADE20K} & \textbf{PASCAL VOC} &
\multicolumn{2}{c}{\textbf{NYU Depth V2}} &
\multicolumn{2}{c}{\textbf{SUN RGB-D}} \\
\cmidrule(lr){5-6}\cmidrule(lr){7-8}
& & & & \textbf{lin. 1} & \textbf{lin. 4} & \textbf{lin. 1} & \textbf{lin. 4} \\
\midrule

OpenCLIP~\citep{Cherti_2023_CVPR} & ViT-G/14 & 39.3 & 71.4 & 0.541 & 0.510 & 0.537 & 0.476 \\
\midrule
MAE~\citep{He_2022_CVPR} & ViT-H/14 & 33.3 & 67.6 & 0.517 & 0.483 & 0.545 & 0.523 \\
DINO~\citep{Caron_2021_ICCV} & ViT-B/8  & 31.8 & 66.4 & 0.555 & 0.539 & 0.553 & 0.541 \\
iBOT~\citep{zhou2022image} & ViT-L/16 & 44.6 & 82.3 & 0.417 & 0.387 & 0.447 & 0.435 \\
\midrule

Low Resolution& ViT-B/14 & 40.6 & 71.2 & 0.423 & 0.408 & 0.463 & 0.439 \\
Medium Resolution & ViT-B/14 & 45.5 & 78.9 & \underline{0.389} & \underline{0.373} & \underline{0.432} & \underline{0.416} \\
High Resolution& ViT-B/14 & \underline{46.1} & \underline{82.5} & 0.394 & 0.380 & 0.445 & 0.426 \\
\shortname{} (Ours) & ViT-B/14 &
\textbf{47.4} &
\textbf{83.5} &
\textbf{0.361}& \textbf{0.358} & \textbf{0.419} & \textbf{0.407}\\
\midrule

\scriptsize \textbf{Rel. Improv} &
- & \scriptsize \textcolor{red}{\textbf{+2.8\%}} & 
\scriptsize \textcolor{red}{\textbf{+1.2\%}} & 
\scriptsize \textcolor{red}{\textbf{+7.2\%}} &
\scriptsize \textcolor{red}{\textbf{+4.0\%}} &
\scriptsize \textcolor{red}{\textbf{+3.0\%}} &
\scriptsize \textcolor{red}{\textbf{+2.2\%}} \\
\bottomrule
\end{tabular}
}
\end{table*}

\begin{figure}[t]
    \centering
    \begin{subfigure}[b]{0.48\textwidth}
        \centering
        \includegraphics[width=\textwidth]{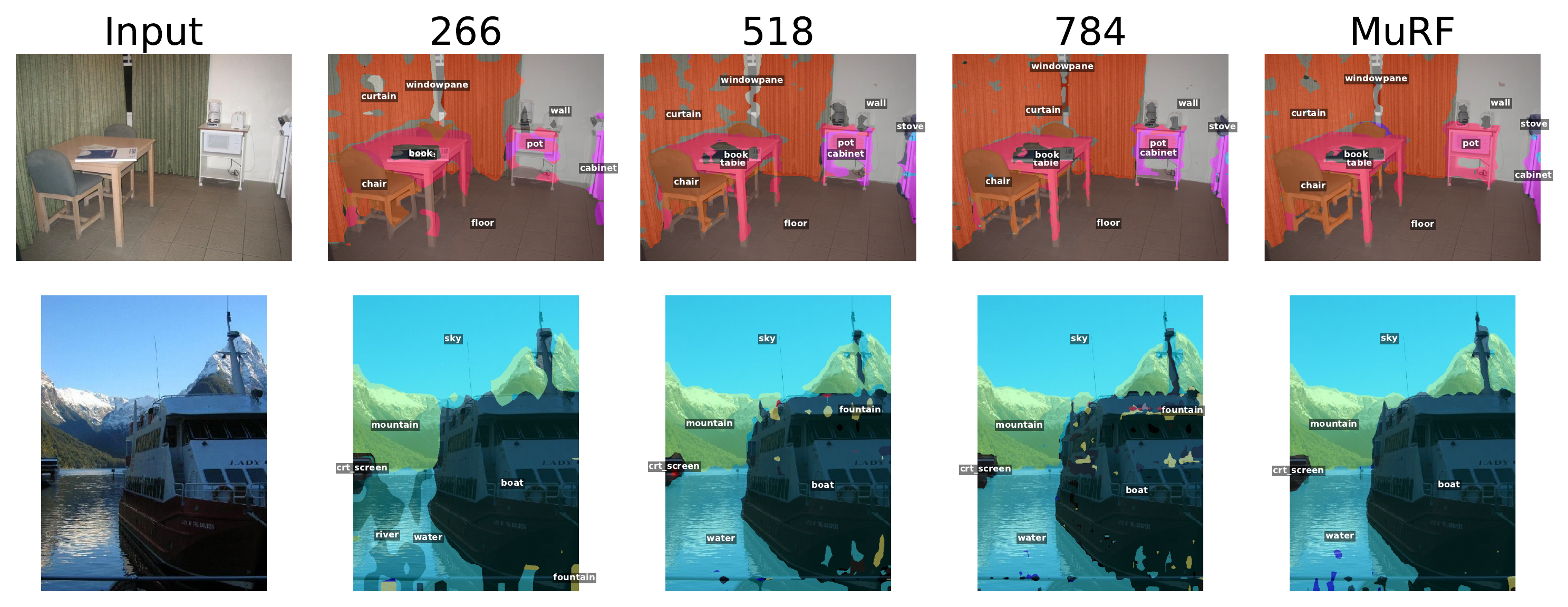}
        \caption{Comparison of semantic segmentation results on ADE20K.}
    \end{subfigure}
    \hfill
    \begin{subfigure}[b]{0.48\textwidth}
        \centering
        \includegraphics[width=\textwidth]{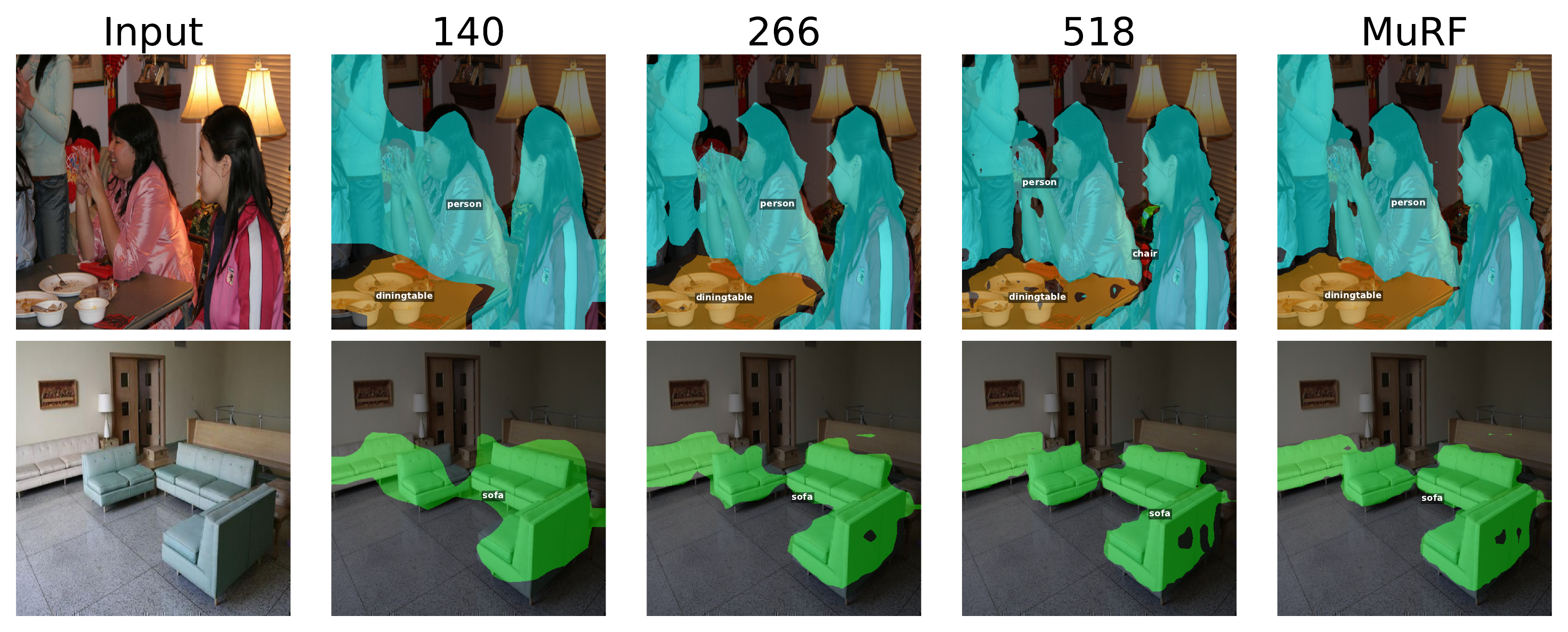}
        \caption{Comparison of semantic segmentation results on PASCAL VOC}
    \end{subfigure}
    \caption{Qualitative comparison of semantic segmentation results on ADE20K (top) and PASCAL VOC (bottom) with different input resolutions. All images are resized to a square shape before being fed into DINOv2, and the subtitle above each image indicates the corresponding input resolution (side length in pixels).}
    \label{fig:seg_results_compare}
\end{figure}

\paragraph{Results.}
Table~\ref{tab:segmentation_depth_combined} presents the main results. Our \shortname{}-based representation provides a significant performance boost over the strong single-scale baseline. Figure~\ref{fig:seg_results_compare} qualitatively illustrates this improvement. This demonstrates that by combining global context from low resolutions and fine details from high resolutions, \shortname{} produces a feature map that is inherently better suited for dense prediction.  For PASCAL VOC segmentation under the linear probing protocol, \shortname{} requires approximately $1.3\times$ the training time of the single-resolution counterpart.

\subsection{Depth Estimation}

\paragraph{Setup.}
We test the in-domain learning capability on \textbf{NYU Depth V2}~\citep{10.1007/978-3-642-33715-4_54} dataset and the transfer learning capability on \textbf{SUN RGB-D} dataset~\citep{7298655}. Following standard protocols, we report performance using Root Mean Squared Error (RMSE, lower is better). Our evaluation employs two linear probing configurations: \textbf{Lin. 1}, which utilizes features from the final transformer layer concatenated with the \texttt{[CLS]} token, and \textbf{Lin. 4}, which follows the same protocol but concatenates tokens from layers $l=\{3, 6, 9, 12\}$. For our \shortname{} method, these feature sets are extracted at multiple resolutions and fused. In both cases, we compare against a baseline frozen DINOv2 backbone with a single-scale input.

\begin{table}[ht]
\centering
\caption{%
Computational Cost and Efficiency Analysis on DINOv2-ViT-B/14. We compare the single-scale baselines against our multi-scale \shortname{} in terms of training latency, downstream head parameter count, and performance (RMSE on NYU Depth V2). \shortname{} achieves the best performance with a reasonable trade-off in computation.}
\label{tab:depth_efficiency}
\resizebox{0.95\columnwidth}{!}{%
\begin{tabular}{lccccc}
\toprule
\textbf{Method} & \textbf{Resolution} & \textbf{Latency (ms/iter)} & \textbf{Head Params (M)} & \textbf{VRAM (GB)} & \textbf{RMSE $\downarrow$} \\
\midrule
Single-Scale & $0.5\times$ & 11.36 & 0.39 & 0.36 & 0.423 \\
Single-Scale & $1.0\times$ & 22.61 & 0.39 & 0.43 & 0.389 \\
Single-Scale & $1.5\times$ & 32.55 & 0.39 & 0.54 & 0.394 \\
\midrule
\textbf{\shortname{} (Ours)} & $\{0.5, 1.0, 1.5\}\times$ & 58.35 & 1.18 & 0.56 & \textbf{0.361} \\
\bottomrule
\end{tabular}%
}
\end{table} 

\begin{figure}[t]
    \centering
    \begin{subfigure}[b]{0.48\textwidth}
        \centering
        \includegraphics[width=\textwidth]{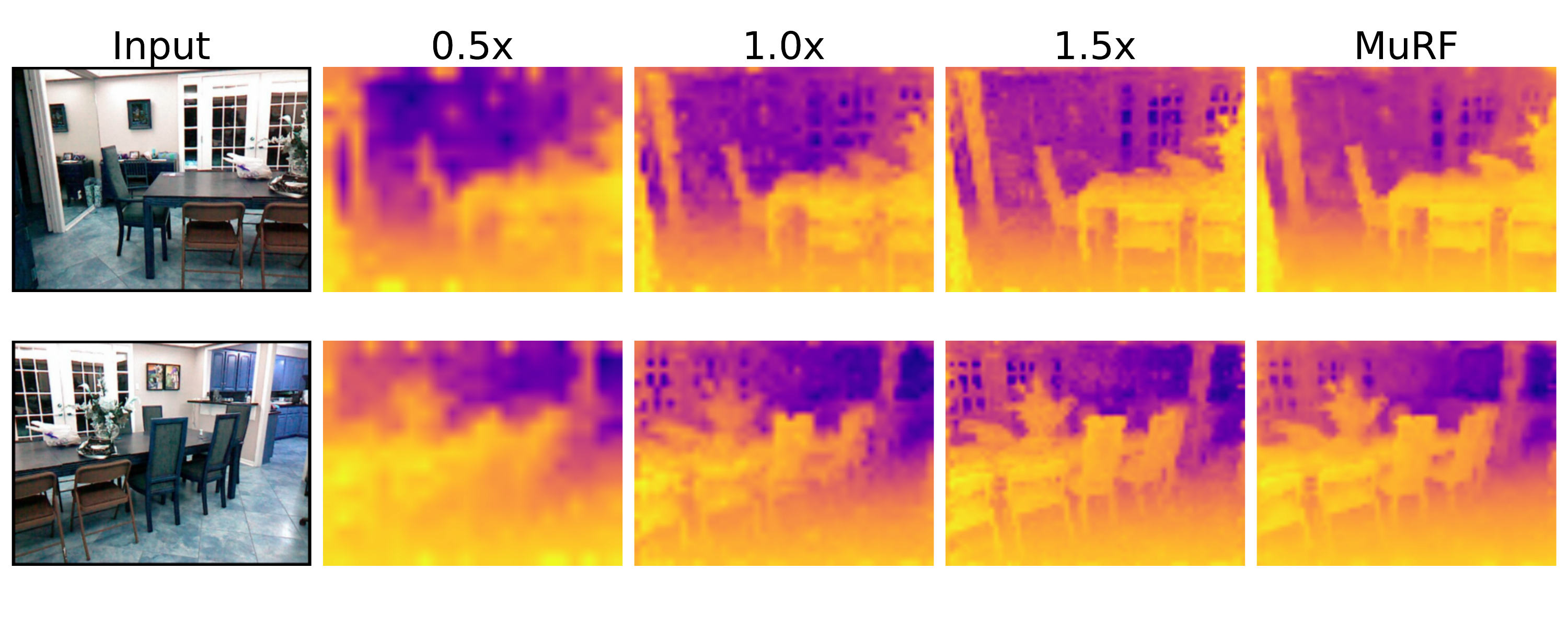}
        \caption{Comparison of depth estimation results on NYUd.}
    \end{subfigure}
    \hfill
    \begin{subfigure}[b]{0.48\textwidth}
        \centering
        \includegraphics[width=\textwidth]{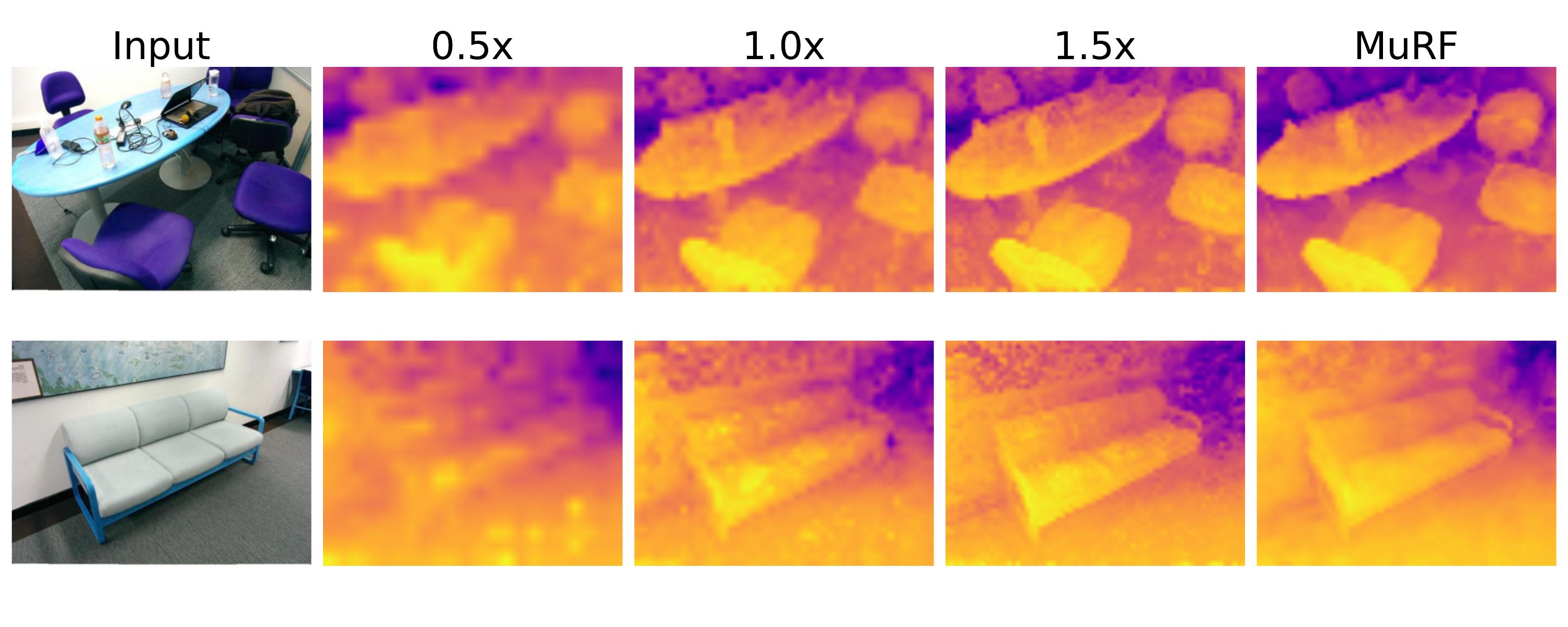}
        \caption{Comparison of depth estimation results on SUN RGB-D}
    \end{subfigure}
    \caption{Qualitative depth estimation results on NYUd (left) and SUN RGB-D (right). We compare single-scale DINOv2 predictions at 0.5×, 1.0×, and 1.5× input resolutions with our \shortname{} fusion. By aggregating multi-resolution features, \shortname{} better preserves global scene structure while sharpening local geometry, producing smoother and more accurate depth maps. Labels 0.X× indicate that the image fed into DINOv2 is resized to 0.X of the original image height and width.}
    \label{fig:depth_results_compare}
\end{figure}

\paragraph{Results.}
As shown in Table~\ref{tab:segmentation_depth_combined} and Figure~\ref{fig:depth_results_compare}, the model using our \shortname{} representation achieves substantially lower error rates. This indicates that the fusion of multi-scale features allows the prediction head to better reason about both the overall scene geometry (from low resolutions) and object boundaries (from high resolutions), leading to more accurate depth predictions. We have also measured the performance impact of MuRF, as demonstrated in Table \ref{tab:depth_efficiency}.

\subsection{Visual Question Answering}

\paragraph{Setup.}
We integrate \shortname{} into an MLLM framework for VQA. The original vision encoder in LLaVA 1.5 is CLIP, which doesn't have natural multi-resolution support. To verify how \shortname{} can support MLLMs, we apply \shortname{} on multiple variants of LLaVA 1.5, where we keep the same training recipe while changing the vision encoder.

In LLaVA's DINOv2 variant, we use a single resolution input ($336 \times 336$) in DINOv2 as a baseline. For \shortname{}, we process resolutions of $224$ and $336$. In LLaVA's SigLIP2 variant, we use a single resolution input ($384 \times 384$) in SigLIP2 as a baseline. For \shortname{}, we process resolutions of $256$ and $384$.

For all experiments, to prevent a computational bottleneck in the LLM, we must strictly avoid increasing the token sequence length (all experiments will have $576$ visual tokens). Instead of appending multi-resolution tokens sequentially, features from the low resolution ($224$ for DINOv2 and $256$ for SigLIP2) are spatially up-sampled and concatenated \textit{patch-wise along the channel dimension} with the high resolution ($336$ for DINOv2 and $384$ for SigLip2) features. The LLM's visual projector then maps these high-channel-dimension tokens back to the standard LLM hidden dimension. \textbf{Crucially, this ensures the number of visual tokens fed into the LLM remains exactly the same as the single-resolution baseline}, incurring zero extra sequence-length computational cost for the LLM.

\paragraph{Results.}
The results in Table~\ref{tab:exp_vqa} show that equipping the MLLM with our \shortname{} representation strongly improves the multimodal understanding capacity regardless of whether DINOv2 or SigLIP2 is used as vision encoder. This suggests that the rich visual context provided by \shortname{} captures both holistic scene understanding and fine-grained details, empowers the language model to answer a wider range of visual questions more accurately. \shortname{} maintains comparable VRAM usage, as well as training and inference latency, relative to the single-resolution baseline. The Single-resolution baseline takes 71 minutes to pretrain, and 270 minutes to finetune, while \shortname{} takes 72 minutes to pretrain, and 274 minutes to finetune.

\begin{table}[]
\centering
\caption{LLaVA-MuRF VQA Performance on MME\citep{fu2025mme}, VLMsAreBiased (Bias)\citep{vo2026vision}, V$*$\citep{Wu_2024_CVPR}, MME RealWorld (MR)\citep{zhang2025mmerealworld}, RealWorld QA (RW)\citep{xai2024grok15}, GQA\citep{Hudson_2019_CVPR}, the total score obtained on four subset of mmbench (MMB)\citep{10.1007/978-3-031-72658-3_13} ("cn\_cc", "cn\_dev", "en\_dev", "ru\_dev"), and the accuracy metric of POPE\citep{li-etal-2023-evaluating}. \textbf{Bold} indicates the best performance.}
\label{tab:exp_vqa}
\small
\setlength{\tabcolsep}{3pt}
\resizebox{1\linewidth}{!}{
\begin{tabular}{llrrrrrrrrr}
\hline
\multirow{2}{*}{\textbf{Vision Encoder}}
 & \multirow{2}{*}{\textbf{Res.}}
 & \multicolumn{2}{c}{\textbf{MME}}
 & \multicolumn{1}{c}{\multirow{2}{*}{\textbf{Bias}}}
 & \multicolumn{1}{c}{\multirow{2}{*}{\textbf{V$*$}}}
 & \multicolumn{1}{c}{\multirow{2}{*}{RW}}
 & \multicolumn{1}{c}{\multirow{2}{*}{MR}}
 & \multicolumn{1}{c}{\multirow{2}{*}{GQA}}
 & \multicolumn{1}{c}{\multirow{2}{*}{MMB}}
 & \multicolumn{1}{c}{\multirow{2}{*}{POPE}} \\
\cline{3-4}
 &  & Percept. & Cogn.
 &  &  &  &  &  &  &  \\
\hline
CLIP (official LLaVA 1.5)
 & 336
 & 1511.4 & 347.1
 & 16.2 & \textbf{50.3} & 56.1 & 26.5 & 62.0
 & 195.4
 & 86.9 \\
\hline
\multirow{2}{*}{DINOv2}
 & 336
 & 1291.6 & 278.6
 & 17.3 & 38.7 & 52.3 & 26.2 & 62.1
 & 172.4
 & 87.1 \\
 & 224+336 (\textbf{Ours})
 & \shortstack[c]{1357.1\\[-1pt]\scriptsize\textcolor{red}{(+65.5)}}
 & \shortstack[c]{366.4\\[-1pt]\scriptsize\textcolor{red}{(+87.8)}}
 & \shortstack[c]{17.7\\[-1pt]\scriptsize\textcolor{red}{(+0.4)}}
 & \shortstack[c]{40.3\\[-1pt]\scriptsize\textcolor{red}{(+1.6)}}
 & \shortstack[c]{53.6\\[-1pt]\scriptsize\textcolor{red}{(+1.3)}}
 & \shortstack[c]{26.1\\[-1pt]\scriptsize(-0.1)}
 & \shortstack[c]{62.4\\[-1pt]\scriptsize\textcolor{red}{(+0.3)}}
 & \shortstack[c]{173.1\\[-1pt]\scriptsize\textcolor{red}{(+0.7)}}
 & \shortstack[c]{87.1\\[-1pt]\scriptsize(0.0)} \\
\hline
\multirow{2}{*}{Clip+DINOv2}
 & 336
 & 1403.4 & 243.9
 & 15.8 & 48.7 & 53.6 & 31.5 & 62.2
 & 194.2
 & 86.4 \\
 & 224+336 (\textbf{Ours})
 & \shortstack[c]{1471.2\\[-1pt]\scriptsize\textcolor{red}{(+67.8)}}
 & \shortstack[c]{281.4\\[-1pt]\scriptsize\textcolor{red}{(+37.5)}}
 & \shortstack[c]{16.3\\[-1pt]\scriptsize\textcolor{red}{(+0.5)}}
 & \shortstack[c]{48.2\\[-1pt]\scriptsize(-0.5)}
 & \shortstack[c]{56.7\\[-1pt]\scriptsize\textcolor{red}{(+3.1)}}
 & \shortstack[c]{31.8\\[-1pt]\scriptsize\textcolor{red}{(+0.3)}}
 & \shortstack[c]{62.9\\[-1pt]\scriptsize\textcolor{red}{(+0.7)}}
 & \shortstack[c]{198.8\\[-1pt]\scriptsize\textcolor{red}{(+4.6)}}
 & \shortstack[c]{\textbf{87.4}\\[-1pt]\scriptsize\textcolor{red}{(+1.0)}} \\
\hline
\multirow{2}{*}{SigLIP2}
 & 384
 & 1529.3 & 355.4
 & 19.4 & 44.0 & 58.2 & 33.1 & 64.1 
 & 211.7
 & 87.1 \\
 & 256+384 (\textbf{Ours})
 & \shortstack[c]{\textbf{1545.7}\\[-1pt]\scriptsize\textcolor{red}{(+16.4)}}
 & \shortstack[c]{\textbf{371.4}\\[-1pt]\scriptsize\textcolor{red}{(+16.0)}}
 & \shortstack[c]{\textbf{19.7}\\[-1pt]\scriptsize\textcolor{red}{(+0.3)}}
 & \shortstack[c]{42.9\\[-1pt]\scriptsize(-1.1)}
 & \shortstack[c]{\textbf{58.4}\\[-1pt]\scriptsize\textcolor{red}{(+0.2)}}
 & \shortstack[c]{\textbf{33.3}\\[-1pt]\scriptsize\textcolor{red}{(+0.2)}}
 & \shortstack[c]{\textbf{64.5}\\[-1pt]\scriptsize\textcolor{red}{(+0.4)}}
 & \shortstack[c]{\textbf{216.9}\\[-1pt]\scriptsize\textcolor{red}{(+5.2)}}
 & \shortstack[c]{86.7\\[-1pt]\scriptsize(-0.4)} \\
\hline
\end{tabular}
}
\end{table}

\subsection{Unsupervised Anomaly Detection}

\paragraph{Setup.}
Finally, we validate \shortname{} in a training-free setting on the \textbf{MVTec AD 2}~\citep{heckler2025mvtec} benchmark, a challenging industrial inspection dataset. We use the pixel-level AU-PRO\textsubscript{0.05} score as the primary metric and compare against state-of-the-art methods like PatchCore~\citep{roth2022towards} and SuperAD~\citep{zhang2025superad}.

For this experiment, we use a set of five scaling factors relative to the original image size: $\{0.3, 0.4, 0.5, 0.6, 0.7\}$. 

\paragraph{Results.}
Table~\ref{tab:anomaly_results} confirms that our method achieves highly competitive, and in several cases state-of-the-art, performance. The ability of \shortname{} to handle anomalies of vastly different scales—from microscopic scratches to large, structural defects—is a direct result of its multi-resolution design, proving its effectiveness even without any parameter tuning.

\begin{table}[ht]
\centering

\captionof{table}{Anomaly detection performance (AU-PRO\textsubscript{0.05} in \%) on the MVTec AD 2 dataset. \shortname{} demonstrates state-of-the-art results on TEST\textsubscript{priv,mix} subset, showcasing its robustness in a challenging training-free scenario. "Training" means if the method involves parameter tuning within a neural network. \textbf{Bold} indicates the best performance, and \underline{underline} indicates the second best.}
\label{tab:anomaly_results}
\resizebox{0.6\linewidth}{!}{
\begin{tabular}{lccc}
\toprule
\small
\textbf{Method} & \textbf{Training?} & \textbf{TEST\textsubscript{priv}} & \textbf{TEST\textsubscript{priv,mix}} \\
\midrule
PatchCore~\citep{roth2022towards} & \xmark & 62.3 & 52.6  \\
SuperAD~\citep{zhang2025superad} & \xmark & 61.2 & 59.3 \\
RoBiS~\citep{li2025robis} & \cmark & \textbf{67.3} & \underline{59.7} \\
\shortname{} (Ours) & \xmark & \underline{66.0} & \textbf{62.3} \scriptsize{\textcolor{red}{$\uparrow$+2.6}}  \\
\bottomrule
\end{tabular}
}

\end{table}

\begin{figure}[t]
  \centering

  \begin{minipage}[t]{0.57\textwidth}
    \centering
    \includegraphics[width=\linewidth]{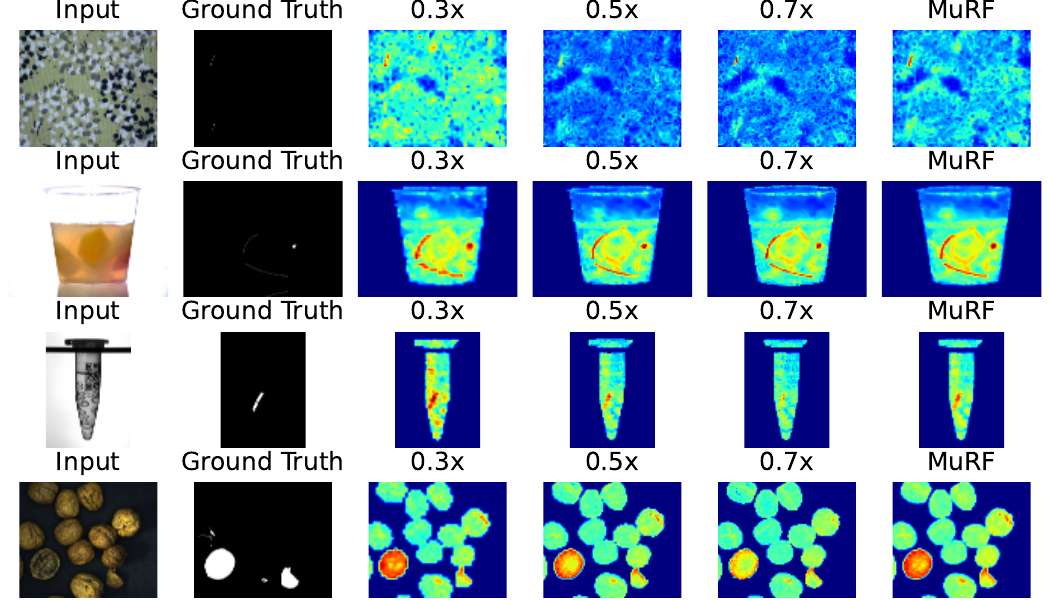}
    \captionof{figure}{The visualization of anomaly detection on MVTec AD 2 TEST\textsubscript{pub} dataset. Our merged result (\shortname{}) successfully combines the robust detection from low-resolution views (e.g., 0.3$\times$ correctly identifies the anomaly's presence but with a coarse mask) and the sharp boundaries from high-resolution views (e.g., 0.7$\times$).}
    \label{fig:results_compare_ad}
  \end{minipage}\hfill
  \begin{minipage}[t]{0.41\textwidth}
    \centering
    \includegraphics[width=\linewidth]{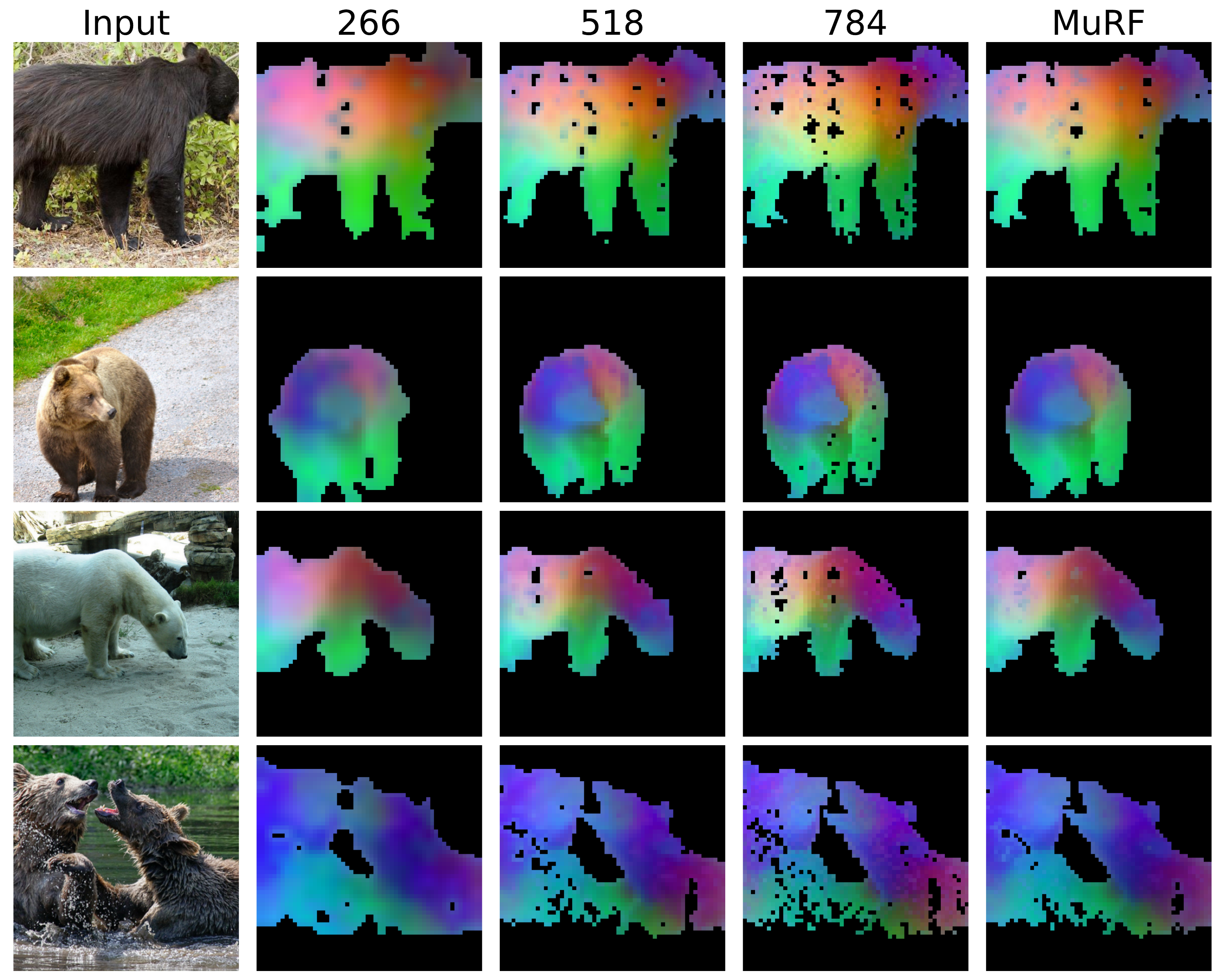}
    \captionof{figure}{The PCA of feature embedding. It can be seen that lower resolution images tend to have better internal representation, but have worse boundaries due to low resolution. Higher resolution images tend to have sharper boundary but have holes in their internal. \shortname{} achieves a good balance that generates high-quality representation in both.}
    \label{fig:results_compare}
  \end{minipage}

\end{figure}

Qualitatively, we sampled some results and presented them in Figure~\ref{fig:results_compare_ad}. The final ``Merged" anomaly maps successfully combine the strengths of different input resolutions. Low-resolution views (e.g., 0.3$\times$) are adept at robustly identifying the presence of an anomaly, though the resulting detection masks are coarse. Conversely, high-resolution views (e.g., 0.7$\times$) provide sharp and precise boundaries for these anomalies. However, those views contain lots of noise or holes inside large anomaly areas (e.g., in walnuts). By fusing these different views, the \shortname-AD method produces more accurate and complete segmentation masks than single-resolution approaches like SuperAD~\citep{zhang2025superad}.

\subsection{Qualitative comparison}

We present the PCA of DINOv2's feature embedding on multiple resolutions, as well as our method's feature embedding.

We have presented the results in Figure~\ref{fig:results_compare}. It's easy to see that high resolution images on the right provide sharper, more accurate boundaries. However, this focus on local detail comes at a cost: the model can lose global context, resulting in ``holes'' within the segmented region, while the low resolution images on the left's internal is smoother. Our proposed approach \shortname{} (on the right) achieved a good balance. 

\subsection{Analysis of Resolution and Feature Concatenation}

To show the benefits of \shortname{}, we conduct a controlled comparison against a standard feature concatenation baseline. To ensure a fair comparison over feature representation, we compare aggregating three distinct resolutions (\shortname{}) against aggregating three distinct encoder layers (Lin. 3).

The results, presented in Table~\ref{tab:linear_comparison}, reveal a complementary relationship between multi-scale and multi-layer features. \shortname{} achieves superior performance on the in-domain NYU Depth V2 dataset (0.361 vs. 0.376), suggesting that scaling effectively captures the fine-grained structural details required for precise metric depth estimation. Conversely, the multi-layer approach (Lin. 3) performs comparably to \shortname{} (0.418 vs. 0.419). This aligns with the intuition that intermediate Transformer layers retain robust, generic semantic abstractions.

Crucially, the best performance is achieved by combining both methodologies. This demonstrates that \shortname{} and multi-layer feature concatenations are not mutually exclusive; rather, they offer orthogonal benefits for complementary performance.

\begin{table}[ht]
\centering

\begin{minipage}[t]{0.66\linewidth}
\centering
\captionof{table}{Linear probing comparison for depth estimation reporting RMSE. We evaluate on NYU Depth V2 (in-domain) and SUN RGB-D (zero-shot). \textbf{Lin. 1} utilizes only the final layer. \textbf{Lin. 3} utilizes layers $\{4, 8, 12\}$. Lower scores indicate better performance. \textbf{Bold} indicates the best performance, \underline{underline} indicates the second best.}
\label{tab:linear_comparison}
\resizebox{\linewidth}{!}{%
\begin{tabular}{l c c c c}
\toprule
\textbf{Method} & \textbf{Resolutions} & \textbf{Layers} & \textbf{NYUd} & \textbf{SUN RGB-D}\\
\midrule
Lin. 1 & $1.0$ & $12$ & 0.389 & 0.432 \\
\shortname{} & $\{0.5, 1.0, 1.5\}$ & $12$ & \underline{0.361} & 0.419 \\
\midrule
Lin. 3 & $0.5$ & $\{4, 8, 12\}$ & 0.412 & 0.443 \\
Lin. 3 & $1.0$ & $\{4, 8, 12\}$ & 0.376 & \underline{0.418} \\
Lin. 3 & $1.5$ & $\{4, 8, 12\}$ & 0.380 & 0.428 \\
\textbf{Lin. 3 + \shortname{}} & $\{0.5, 1.0, 1.5\}$ & $\{4, 8, 12\}$ & \textbf{0.357} & \textbf{0.409} \\
\bottomrule
\end{tabular}%
}
\end{minipage}
\hfill
\hfill
\begin{minipage}[t]{0.32\linewidth}
\centering
\captionof{table}{Semantic segmentation performance in mIoU (\%) when using SigLIP2 as backbone. \textbf{Bold} indicates the best performance, \underline{underline} indicates the second best.}
\label{tab:siglip2_segmentation_results}
\begin{tabular}{lc}
\toprule
\textbf{Resolution} & \textbf{ADE20K$\uparrow$}\\
\midrule
256 & 32.04 \\
512 & \underline{35.27} \\
768 & 34.46\\
\midrule
\shortname{} (Ours) & \textbf{37.10} \scriptsize{\textcolor{red}{$\uparrow$+1.83}} \\
\bottomrule
\end{tabular}
\end{minipage}

\end{table}

\subsection{Ablation Study}

We studied how the number of resolutions fused into \shortname{} has an impact on the performance on two downstream tasks: depth estimation and anomaly detection.

\subsubsection{Depth Estimation}
To isolate the contribution of each scale, we conduct an ablation study on the depth estimation task. We evaluate performance on NYUd using the \textbf{Lin. 1} protocol (reporting RMSE) while varying the specific resolutions used in our fusion. The results in Table~\ref{tab:abl_mrf_depth} compare single-scale baselines against two-scale and our full three-scale \shortname{} method.

\begin{table*}[t]
\centering

\begin{minipage}[t]{0.49\textwidth}
\centering
\setlength{\tabcolsep}{8pt}
\caption{\textbf{\shortname{} ablation results for depth estimation.} We compare single-scale baselines and two-scale fusions against the full three-scale \shortname{}. Performance is measured in RMSE (lower is better). The performance drops (indicated in \textcolor{red}{red}) are relative to the Full \shortname{} setting.}
\label{tab:abl_mrf_depth}
\resizebox{\linewidth}{!}{%
\begin{tabular}{l ccc l}
\toprule
& \multicolumn{3}{c}{Resolutions} & \\
\cmidrule(lr){2-4}
Setting & 0.5$\times$ & 1.0$\times$ & 1.5$\times$ & RMSE $\downarrow$ \\
\midrule
\multicolumn{5}{l}{\textit{Single Resolution}} \\ \midrule
1 & \cmark & & & 0.423 \scriptsize{\textcolor{red}{$\uparrow$0.062}} \\
2 & & \cmark & & 0.389 \scriptsize{\textcolor{red}{$\uparrow$0.027}} \\
3 & & & \cmark & 0.394 \scriptsize{\textcolor{red}{$\uparrow$0.033}} \\
\midrule
\multicolumn{5}{l}{\textit{Fused Resolutions}} \\ \midrule
4 & \cmark & \cmark & & 0.373 \scriptsize{\textcolor{red}{$\uparrow$0.011}} \\
5 & \cmark & & \cmark & 0.366 \scriptsize{\textcolor{red}{$\uparrow$0.005}} \\
6 & & \cmark & \cmark & 0.368 \scriptsize{\textcolor{red}{$\uparrow$0.007}} \\
\midrule
\textbf{\shortname{}} & \textbf{\cmark} & \textbf{\cmark} & \textbf{\cmark} & \textbf{0.361} \\
\bottomrule
\end{tabular}%
}
\end{minipage}
\hfill
\begin{minipage}[t]{0.49\textwidth}
\centering
\setlength{\tabcolsep}{5pt}
\caption{\textbf{Ablation study on Multi-Resolution Settings.} Performance measured by AU-PRO\textsubscript{0.05} on MV Tec AD 2 TEST\textsubscript{pub} dataset is reported. We analyze the impact of different resolution combinations. The performance drops are relative to the Full \shortname{} setting.}
\label{tab:abl_ad}
\resizebox{\linewidth}{!}{%
\begin{tabular}{lcccccl}
\toprule
& \multicolumn{5}{c}{Resolutions} & \\
\cmidrule(lr){2-6}
Setting & 0.3$\times$ & 0.4$\times$ & 0.5$\times$ & 0.6$\times$ & 0.7$\times$ & AU-PRO\textsubscript{0.05} $\uparrow$ \\
\midrule
\multicolumn{7}{l}{\textit{Single Resolution}} \\ \hline
1 & \cmark & & & & & 52.29 \scriptsize\textcolor{red}{$\downarrow$5.03} \\
2 & & \cmark & & & & 53.42 \scriptsize\textcolor{red}{$\downarrow$3.90} \\
3 & & & \cmark & & & 55.39 \scriptsize\textcolor{red}{$\downarrow$1.93} \\
4 & & & & \cmark & & 55.23 \scriptsize\textcolor{red}{$\downarrow$2.09} \\
5 & & & & & \cmark & 54.87 \scriptsize\textcolor{red}{$\downarrow$2.45} \\
\midrule
\multicolumn{7}{l}{\textit{Fused Resolutions}} \\ \midrule
6 & \cmark & & \cmark & & \cmark & 56.60 \scriptsize\textcolor{red}{$\downarrow$0.72} \\
7 & \cmark & \cmark & & \cmark & \cmark & 57.29 \scriptsize\textcolor{red}{$\downarrow$0.03} \\
8 & & & \cmark & \cmark & \cmark & 57.05 \scriptsize\textcolor{red}{$\downarrow$0.27} \\
\midrule
\textbf{\shortname{}} & \cmark & \cmark & \cmark & \cmark & \cmark & \textbf{57.32} \\
\bottomrule
\end{tabular}%
}
\end{minipage}

\end{table*}

All three single-scale settings are clearly suboptimal compared to any multi-scale fusion: the best single-scale baseline (1.0×) achieves an RMSE of 0.389, whereas all two-scale variants already reduce the error to the 0.373–0.366 range, and the full \shortname{} configuration further improves it to 0.361. While the medium and high resolutions (1.0× and 1.5×) individually outperform the low-resolution setting (0.5×), including the coarsest view in the fusion remains beneficial. Comparing the higher-resolution pair (1.0×–1.5×, RMSE 0.366) to full \shortname{} (0.5×–1.0×–1.5×, RMSE 0.361) shows that adding the low-resolution view still yields a measurable gain, even when two relatively strong scales are already present.

This pattern supports our hypothesis that low-resolution inputs provide complementary global geometric context that cannot be fully compensated by higher-resolution views alone. Overall, performance improves monotonically as we increase the number of fused resolutions, with diminishing but consistent returns, confirming that \shortname's benefits stem from genuine multi-scale complementarity rather than from any single particularly “good” resolution.

\subsubsection{Semantic Segmentation}

We performed experiments on SigLIP2-Base-Patch16-NaFlex\citep{tschannen2025siglip} in addition to DINOv2-Base to verify the applicability of our proposed approach. Since SigLIP2 uses patch size of 16 instead of 14, we used the following combination of resolutions:  $S_{\text{res}} = \{256, 512, 768\}$.  Results are presented in Table \ref{tab:siglip2_segmentation_results}.

\subsubsection{Anomaly Detection}

We evaluated different combinations of resolutions to test how the number of resolutions and how the choice of different resolutions have an impact on performance. We used AU-PRO\textsubscript{0.05} on the MV Tec AD 2 TEST\textsubscript{pub} dataset to evaluate the performance.

As with depth estimation, single-scale models are consistently weaker than their multi-scale counterparts, shown in Table~\ref{tab:abl_ad}. The best single resolution is $0.5\times$ (55.39 $\mathrm{AU}\text{-}\mathrm{PRO}_{0.05}$), while both lower ($0.3\times$, 52.29) and higher ($0.7\times$, 54.87) scales alone underperform, indicating that extremely coarse views lose too much detail and extremely fine views become overly sensitive to local noise. Fusing multiple nearby resolutions mitigates these issues: a three-scale fusion of $\{0.3, 0.5, 0.7\}$ already improves performance to 56.60, and adding additional intermediate scales ($\{0.3, 0.4, 0.6, 0.7\}$) further raises it to 57.29, nearly matching the full five-scale \shortname{} (57.32). Notably, different multi-scale subsets with overlapping ranges (e.g., $\{0.5, 0.6, 0.7\}$) perform similarly well, suggesting that what matters most is covering a spectrum of coarse-to-fine views rather than any particular ``magic'' resolution.

These trends align with our qualitative findings in Fig.~\ref{fig:results_compare_ad}: coarse resolutions excel at reliably localizing anomalous regions but yield coarse masks, while finer resolutions sharpen boundaries but may miss parts of large defects. The progressive improvement from single- to multi-resolution settings confirms that \shortname{}'s anomaly detection gains arise from combining these complementary behaviors across scales, rather than from a specific hand-picked input size.

\section{Conclusion}

In this work, we introduced \methodsfullname~(\methodshortname{}), a simple yet powerful inference-time strategy to enhance Vision Foundation Models representations. By constructing a feature pyramid from the input space and fusing the resulting representations, \shortname{} effectively synergizes the global context from low-resolution views with the fine-grained detail from high-resolution ones. Our extensive experiments demonstrate that this single, unified approach yields consistent and significant performance gains across a diverse set of fundamental vision tasks, including dense prediction, multimodal reasoning, and unsupervised anomaly detection. These results establish multi-resolution aggregation not as a task-specific trick, but as a general principle for unlocking the full potential of pre-trained visual encoders.

\section{Acknowledgment}

This work was supported in part by NSF IIS2404180,  KLA,  and the Institute of Information \& Communications Technology Planning \& Evaluation (IITP) grants funded by the Korea government (MSIT) (No. 2022-0-00871, Development of AI Autonomy and Knowledge Enhancement for AI Agent Collaboration), (No. RS-2022-00187238, Development of Large Korean Language Model Technology for Efficient Pretraining), and (No. RS-2025-2543949. Environment-Aware and Domain-Adaptive Multimodal Embodied AI for Real-World Interaction).

\clearpage
\bibliography{main}
\newpage
\appendix
\section{Implementation Details}

\subsection{Semantic Segmentation}
For semantic segmentation, we follow a consistent strategy with the DINOv2 segmentation protocol. We evaluate our method on both the \textit{ADE20K} dataset\footnote{The ADE20K dataset can be obtained at \url{https://data.csail.mit.edu/places/ADEchallenge/ADEChallengeData2016.zip}} and the \textit{PASCAL VOC 2012} dataset. For \shortname{} , we selected $S_{\text{res}} = \{266, 518, 784\}$ for \textit{ADE20K} dataset, and $S_{\text{res}} = \{140, 266, 518\}$ for \textit{PASCAL VOC 2012} dataset since images in \textit{PASCAL VOC 2012} dataset generally has lower resolution.

All experiments are performed using a frozen DINOv2-Base as backbone. The objective function is the standard pixel-wise Cross-Entropy loss. The segmentation head is optimized using the AdamW optimizer with an initial learning rate $\eta = 1 \times 10^{-3}$, weight decay $\lambda_{wd} = 1 \times 10^{-4}$, and momentum parameters $\beta_1 = 0.9, \beta_2 = 0.999$.

For \textit{ADE20K} dataset training, we applied a joint transformation
$\mathcal{T}(\mathbf{x}, \mathbf{y})$ to each RGB input image
$\mathbf{x}$ and its corresponding segmentation mask $\mathbf{y}$. When
data augmentation is enabled, $\mathcal{T}$ is given by the following
composition of operations. We first rescale the image and mask with an isotropic resize operator $\mathcal{R}$ such that the smaller spatial dimension of the image does not exceed $784$ pixels, while preserving the aspect ratio. From the resized image we sample a random square crop $\mathbf{x}_{\text{crop}} \in \mathbb{R}^{c \times c}$ of size $784 \times 784$, and extract the corresponding crop of the mask $\mathbf{y}_{\text{crop}}$. With probability $p = 0.5$ we apply a horizontal flip to both the cropped image and mask. The (possibly flipped) crop is then normalized channel-wise using a fixed mean $\boldsymbol{\mu} \in \mathbb{R}^3$ and standard deviation $\boldsymbol{\sigma} \in \mathbb{R}^3$, $\boldsymbol{\mu} = \left(\frac{123.675}{255}, \frac{116.280}{255}, \frac{103.530}{255}\right)$, $\boldsymbol{\sigma} = \left(\frac{58.395}{255}, \frac{57.120}{255}, \frac{57.375}{255}\right)$ and each RGB channel is transformed as $\tilde{\mathbf{x}} = \frac{\mathbf{x} - \boldsymbol{\mu}}{\boldsymbol{\sigma}}$. We train for a total of $50$ epochs with a batch size of $32$.

For \textit{ADE20K} dataset testing, we normalized the image and resize then to the largest possible scale ($784$). We directly evaluated the output of the model without applying sliding window. The model is evaluated on the original \textit{ADE20K} validation set.

For \textit{PASCAL VOC 2012} dataset training, we utilize the augmented training set combined with the standard training set, which consists of $12,031$ images. We apply a composition of geometric and photometric transformations $\mathcal{T}(\mathbf{x})$ to input images $\mathbf{x}$. First, images are resized such that their shortest spatial dimension is 512 pixels, followed by a random multi-scale resizing operation $\mathcal{S}(\mathbf{x}, s)$ where the scale ratio $s$ is sampled uniformly from $[0.5, 2.0]$. From the rescaled image, we extract a random content-aware crop $\mathbf{x}_{crop} \in \mathbb{R}^{512 \times 512}$. random sampling strategy. To prevent class imbalance during training, this strategy rejects crops where a single valid category occupies more than 75\% of the spatial area. We next apply a horizontal flip with probability $p=0.5$, followed by photometric distortions $\mathcal{P}(\cdot)$. Specifically, we sequentially apply random brightness shifts ($\delta \in [-32, 32]$), contrast and saturation scaling (factor $\in [0.5, 1.5]$), and hue shifts ($\delta \in [-18, 18]$), where each transformation is applied with a probability of $p=0.5$. Finally, inputs are normalized using the ImageNet mean $\mu$ and standard deviation $\sigma$. We train for a total of $40,000$ iterations with a batch size of 16. We employ a sequential learning rate schedule: a linear warm-up is applied for the first $1,500$ iterations (starting from $\eta = 1 \times 10^{-6}$), followed by a polynomial decay schedule with power $p=1.0$ for the remainder of the training.

For \textit{PASCAL VOC 2012} dataset testing, we employ a sliding window evaluation strategy to handle varying image aspect ratios. We use a window size of $512 \times 512$ with a stride of $341 \times 341$ to generate the final segmentation maps. The model is evaluated on the original \textit{PASCAL VOC 2012} validation set.

The code for semantic segmentation experiments has been uploaded. The semantic segmentation experiment was conducted on a single server equipped with a RTX A6000 GPU.

\subsection{Depth Estimation}

We adopt a training strategy consistent with the DINOv2 depth estimation protocol. The model is trained on the \textit{NYU Depth V2} dataset ($\mathcal{D}_{\text{train}}$) and evaluated in a zero-shot setting on \textit{SUN RGB-D} ($\mathcal{D}_{\text{test}}$). we selected $S_{\text{res}} = \{0.5s, 1.0s, 1.5s\}$, where $s$ is the original resolution. Training is performed using mixed-precision floating point arithmetic. We minimize the objective using the AdamW optimizer with an initial learning rate $\eta = 1 \times 10^{-4}$, momentum parameters $\beta_1 = 0.9, \beta_2 = 0.999$, and weight decay $\lambda_{\textrm{wd}} = 0.01$. The training spans $38,400$ iterations with a batch size of 2. We employ a cosine annealing schedule, decaying $\eta$ to $\eta_{\text{min}} = 1 \times 10^{-6}$, and a linear warm-up period of $1,000$ iterations is applied to stabilize early-stage convergence.

We test two feature extraction configurations on the frozen DINOv2-Base encoder $\Phi$: a single-layer setup (Lin. 1) and a multi-layer fusion setup (Lin. 4). In the Lin. 1 configuration, features are extracted solely from the final transformer layer. In the Lin. 4 configuration, we designate a set of intermediate layers $S_{\text{layer}} = \{3, 6, 9, 12\}$. For both configurations, for every utilized layer $l$, we concatenate the layer-specific global classification token $[\text{CLS}]_l$ to each spatial patch token. In the multi-layer case, the final representation is obtained by concatenating these upsampled features across all $l \in S_{\text{layer}}$ along the channel dimension before passing them to the decoding head.

Following the methodology of \cite{Bhat_2021_CVPR}, the decoding head treats depth estimation as a per-pixel classification task. We divide the depth prediction range into 256 uniformly distributed bins over a depth range of $[1\times10^{-3}, 10]$.

\begin{figure}[th]
    \centering
    \begin{subfigure}[b]{0.7\textwidth}
        \centering
        \includegraphics[width=\textwidth]{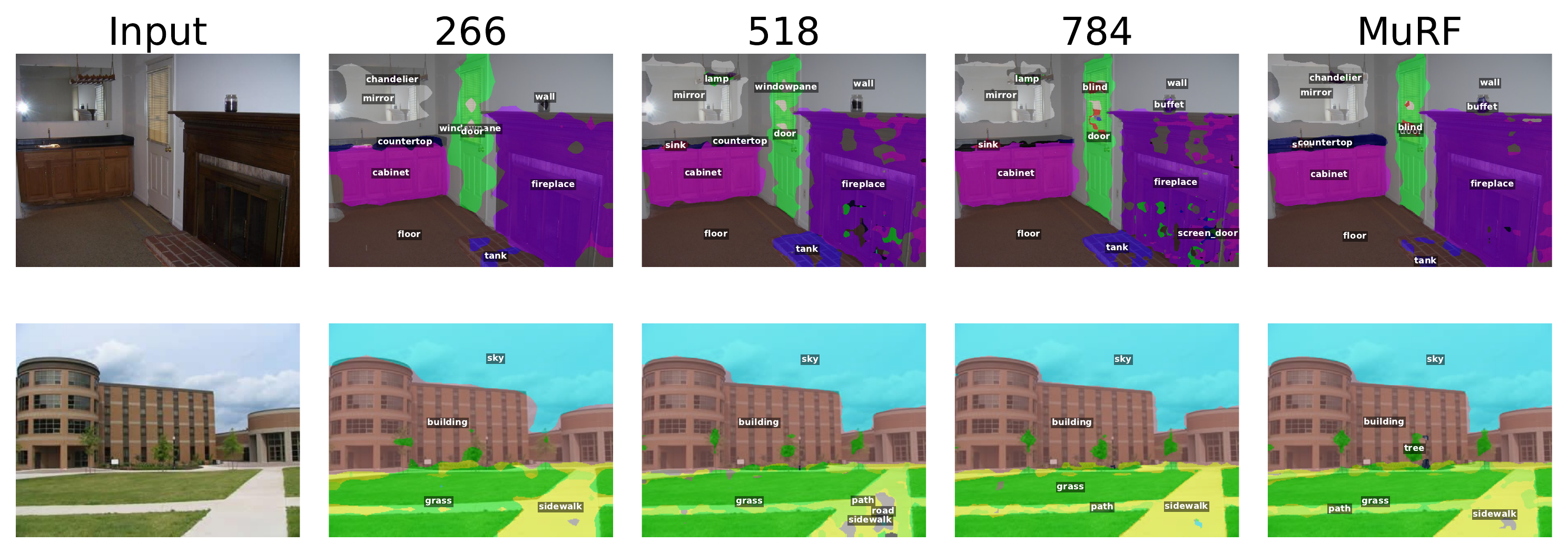}
        \caption{Comparison of semantic segmentation results on ADE20K.}
    \end{subfigure}
    \hfill
    \begin{subfigure}[b]{0.7\textwidth}
        \centering
        \includegraphics[width=\textwidth]{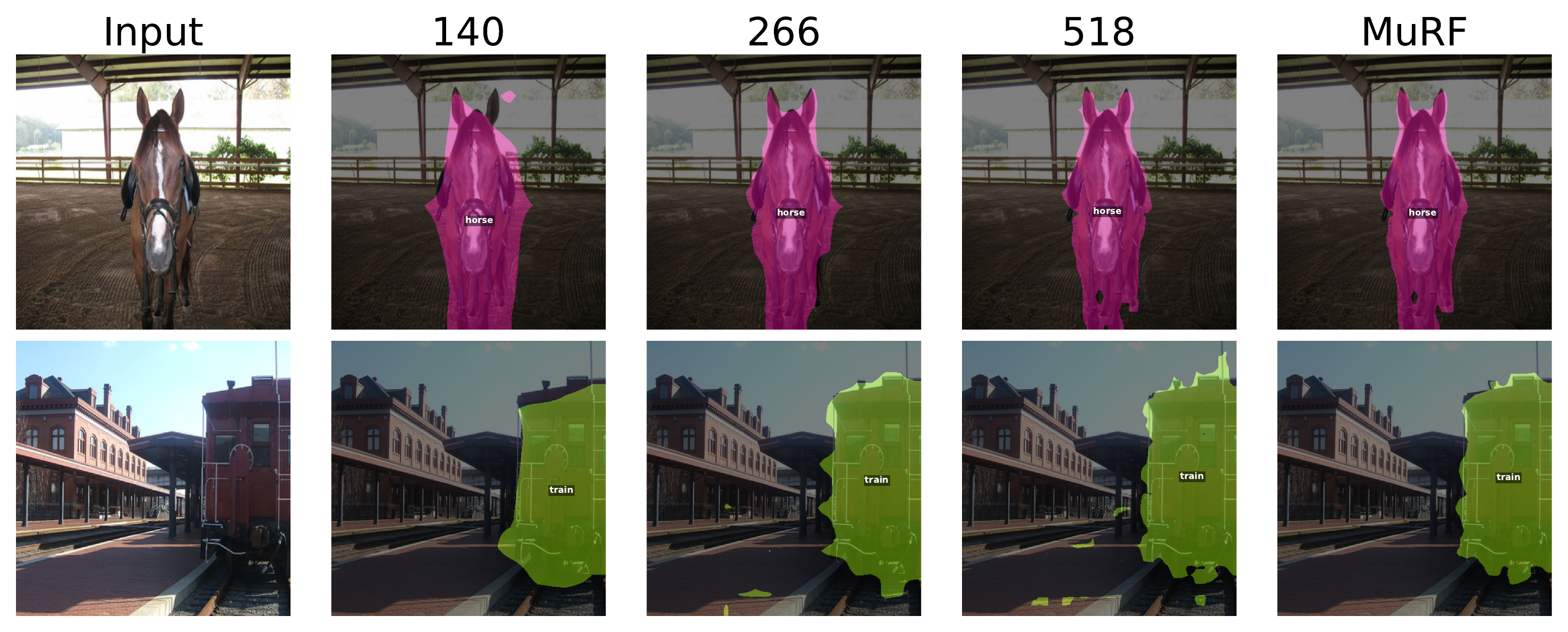}
        \caption{Comparison of semantic segmentation results on PASCAL VOC}
    \end{subfigure}
    \caption{Additional segmentation visualizations from ADE20K and PASCAL VOC.}
    \label{fig:seg_results_compare_supp}
\end{figure}

\begin{figure}[th]
    \centering
    \begin{subfigure}[b]{0.7\textwidth}
        \centering
        \includegraphics[width=\textwidth]{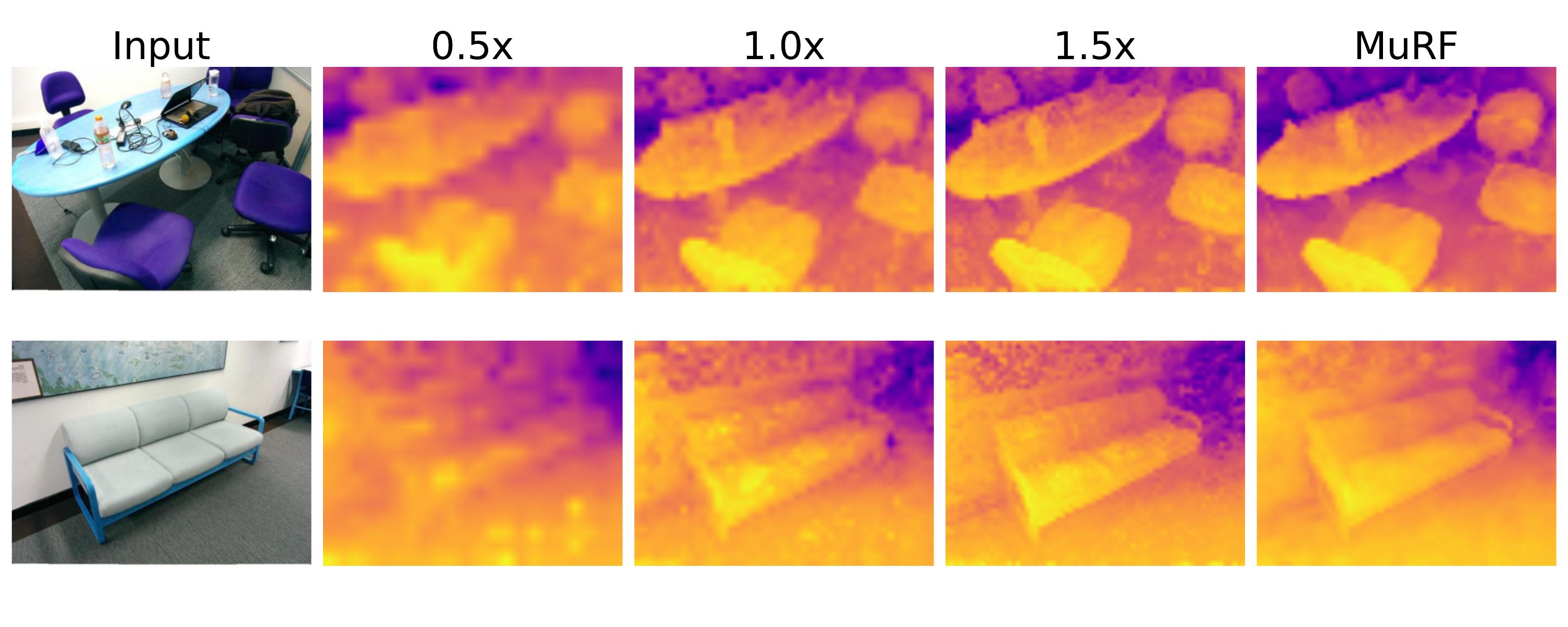}
        \caption{ Additional comparison of depth estimation results on NYUd.}
    \end{subfigure}
    \hfill
    \begin{subfigure}[b]{0.7\textwidth}
        \centering
        \includegraphics[width=\textwidth]{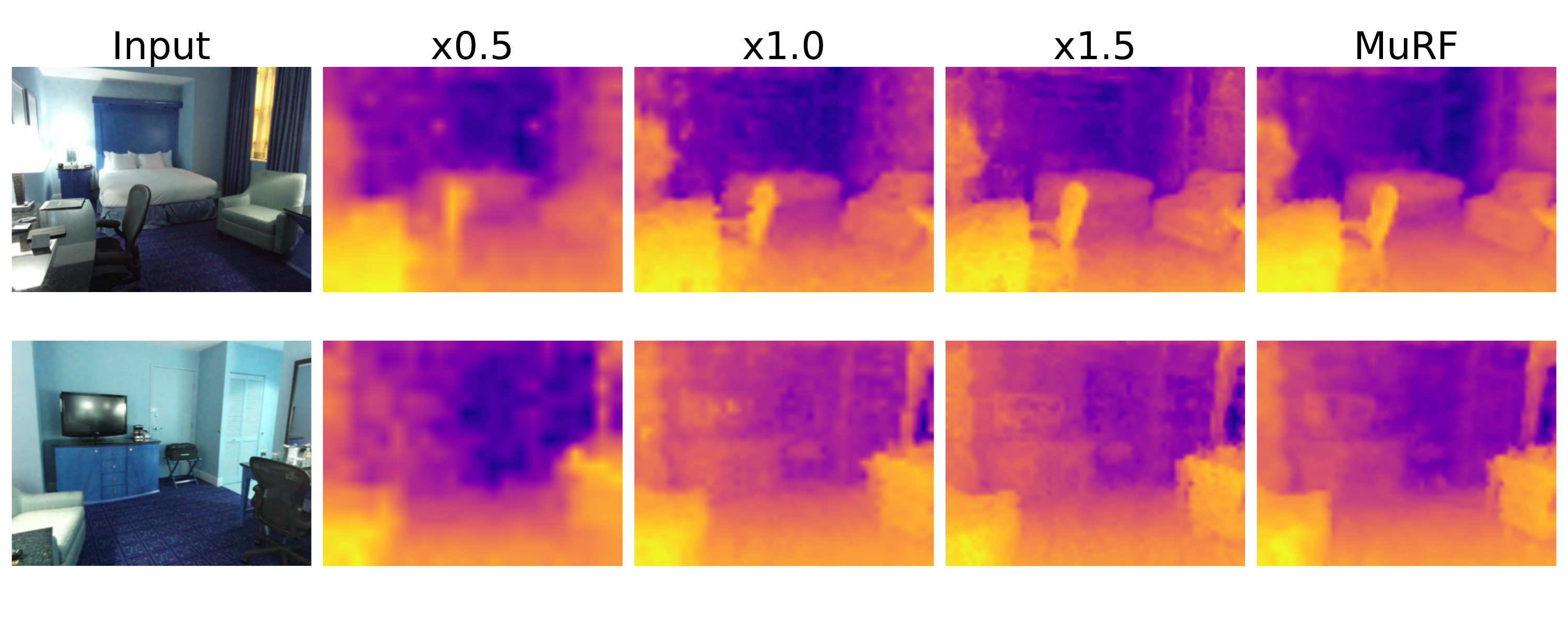}
        \caption{Additional comparison of depth estimation results on SUN RGB-D.}
    \end{subfigure}
    \caption{Additional visualizations of depth estimation results on NYUd and SUN RGB-D.}
    \label{fig:depth_results_compare_supp}
\end{figure}

We apply a composition of geometric and photometric transformations $\mathcal{T}(\mathbf{x})$ to input images $\mathbf{x} \in \mathbb{R}^{H \times W \times C}$. We first apply the standard NYU Eigen crop \citep{NIPS2014_91c56ce4} to remove invalid border regions. We then apply a random rotation $R(\theta)$ with zero-padding, where $\theta \sim \mathcal{U}(-2.5^\circ, 2.5^\circ)$ with probability $p=0.5$, followed by a random horizontal flip with $p=0.5$. Subsequently, we extract a random crop $\mathbf{x}_\text{crop}\in \mathbb{R}^{416 \times 544}$ from the transformed image. We also employ photometric distortions with probability $p=0.5$, specifically color jittering defined by random adjustments to gamma $\gamma \in [0.9, 1.1]$, brightness $\beta \in [0.75, 1.25]$, and per-channel RGB scaling $s_c \in [0.9, 1.1]$. Finally, all inputs are normalized using the ImageNet mean $\mu$ and standard deviation $\sigma$.

The network parameters are optimized by minimizing a composite loss function $\mathcal{L}_{\text{total}}$, consisting of a scale-invariant log loss $\mathcal{L}_{\text{depth}}$ with a warm-up schedule~\citep{Bhat_2021_CVPR}, and a multi-scale gradient matching term $\mathcal{L}_{\text{grad}}$:
\begin{equation}
    \mathcal{L}_{\text{total}} = \lambda_{\text{depth}} \mathcal{L}_{\text{depth}}(\hat{\mathbf{d}}, \mathbf{d}) + \lambda_{\text{grad}} \mathcal{L}_{\text{grad}}(\hat{\mathbf{d}}, \mathbf{d})
    \label{eq:loss}
\end{equation}
where $\hat{\mathbf{d}}$ and $\mathbf{d}$ denote the predicted and ground-truth depth maps, respectively. The term $\mathcal{L}_{\text{grad}}$ enforces gradient consistency across four spatial scales to preserve structural details. We set the balancing coefficients to $\lambda_{\text{depth}}=1.0$ and $\lambda_{\text{grad}}=0.5$.

During evaluation, we process inputs at their native resolution ($480 \times 640$). To further improve predictive stability, we employ Test-Time Augmentation (TTA). The final prediction $\hat{\mathbf{d}}_{\text{final}}$ is computed as the average of the direct prediction and the inverse-transformed prediction of the horizontally flipped input:
\begin{equation}
    \hat{\mathbf{d}}_{\text{final}} = \frac{1}{2} \left( \mathcal{M}(\mathbf{x}) + \operatorname{flip}^{-1}(\mathcal{M}(\operatorname{flip}(\mathbf{x}))) \right)
\end{equation}

The depth estimation experiment was conducted on a single server equipped with a RTX A6000 GPU.

\subsection{LLaVA-style MLLM training}

To verify how \shortname{} can support MLLMs, we replace the vision encoder of LLaVA 1.5 with DINOv2 and SigLIP2. 

For the DINOv2 variant's baseline, we resize all images into squares of $336\times336$, and put them into DINOv2. For \shortname{}, we used $S_{\text{res}} = \{224, 336\}$, and after that, we resize the feature embedding generated by images of size $224\times224$ to $24\times24$ (the same as the feature embedding generated by images of size $336\times336$) using bilinear interpolation. 

For the SigLIP2 variant's baseline, we resize all images into squares of $384\times384$, and put them into SigLIP2. We choose $384$ because SigLIP2 has a patch size of 16 and $384\times384$ image can result in 576 tokens, which is the same for the DINOv2 variant and the same with the original Clip variant. For \shortname{}, we used $S_{\text{res}} = \{256, 384\}$, and after that, we resize the feature embedding generated by images of size $256\times256$ to $24\times24$ (the same as the feature embedding generated by images of size $384\times384$) using bilinear interpolation. 

After that, we perform the token-wise concatenation for all two features embedding, so the length per visual token becomes two times longer but the number of total number of visual token remains the same. We changed the input dimension of the two-layer multi-projection layer accordingly, but keep the output dimension the same. Therefore, the number of tokens inputted into the LLM is the same.

We followed the LLaVA 1.5 two-stage training pipeline with the same hyperparameter.

All experiments were conducted on a single server equipped with eight NVIDIA H100 GPUs.

\subsection{Anomaly Detection}

For anomaly detection, we followed the Embedding-based Anomaly Detection Paradigm as \cite{roth2022towards, zhang2025superad}. 

Specifically, for a given input image $\mathbf{x} \in \mathbb{R}^{H\times W\times C}$, we first generate a set of resized versions $\{\mathbf{x}_s\}_{s \in S_{\text{res}}}$, where $S_{\text{res}}$ is a predefined set of scaling factors. Each version $\mathbf{x}_s$ is then processed by a frozen DINOv2-Base encoder, which we denote as $\Phi$.

To further enrich the representation with hierarchical semantic information, we also extract features from multiple intermediate layers of the encoder, $S_{\text{layer}}$. This process yields a comprehensive collection of patch-level feature maps $\{\mathcal{F}_{l,s} \mid l \in S_{\text{layer}}, s \in S_{\text{res}}\}$, where each feature map is defined as:
\begin{equation}
    \mathcal{F}_{l,s} = \Phi_l(\mathbf{x}_s) \in \mathbb{R}^{H_{l,s} \times W_{l,s} \times d}
\end{equation}
Here, $\Phi_l$ represents the feature output at layer $l$, $(H_{l,s}, W_{l,s})$ are the spatial dimensions of the feature map for scale $s$, and $d$ is the feature dimension. Each feature map $\mathcal{F}_{l,s}$ thus captures a unique combination of spatial scale and semantic level.

At inference time, for a given test image $\mathbf{x}^*$, we extract its feature maps $\{\mathcal{F}_{l,s}^*\}$. For each feature map $\mathcal{F}_{l,s}^*$, we compute a corresponding anomaly score map $\hat{S}_{l,s}$ by scoring each of its feature vectors. The score for a feature vector $\mathbf{f}^* \in \mathcal{F}_{l,s}^*$ is its $L_2$ distance to the nearest neighbor in the dedicated memory bank $\mathcal{M}_{l,s}$:
\begin{equation}
    S(\mathbf{f}^*) = \min_{\mathbf{f} \in \mathcal{M}_{l,s}} \| \mathbf{f}^* - \mathbf{f} \|_2
    \label{eq:scoring}
\end{equation}

Finally, to produce the single output anomaly score map $\hat{\mathbf{S}}$, we fuse all individual score maps $\{\hat{S}_{l,s}\}$. Each map is first up-sampled to the original image dimensions $(H, W)$ via bilinear interpolation, and then aggregated through element-wise averaging:
\begin{equation}
    \hat{\mathbf{S}} = \frac{1}{|S_{\text{layer}}||S_{\text{res}}|} \sum_{l \in S_{\text{layer}}} \sum_{s \in S_{\text{res}}} \text{Upsample}(\hat{S}_{l,s})
    \label{eq:fusion}
\end{equation}

To streamline the process in different resolution through a standard scaling factor, we preprocess images in different categories of the MVTec AD 2 dataset by resizing them to comparable total number of pixels, as shown in Table~\ref{tab:resolution_conversion}.

\begin{table}[h]
\centering
\caption{Resolution conversion table for MVTec AD v2 categories to standardize the total number of pixels for our experiments.} 
\label{tab:resolution_conversion}
\resizebox{0.75\linewidth}{!}{
\begin{tabular}{lcccc}
 \toprule
Object & Original Resolution  & \# of Pixels  & New Resolution & New \# of Pixels \\
\midrule
Can & 1024x2232 & 2,285,568 & 1536x3348 & 5,142,528 \\
Fruit Jelly & 1520x2100 & 3,192,000 & 1900x2625 & 4,987,500 \\
Vial & 1900x1400 & 2,660,000 & 2470x1820 & 4,495,400 \\
\multicolumn{5}{l}{\textit{(... other categories with no change are omitted for brevity ...)}} \\
\bottomrule
\end{tabular}
}
\end{table}

\begin{figure}[t]
    \centering
    \includegraphics[width=0.48\textwidth]{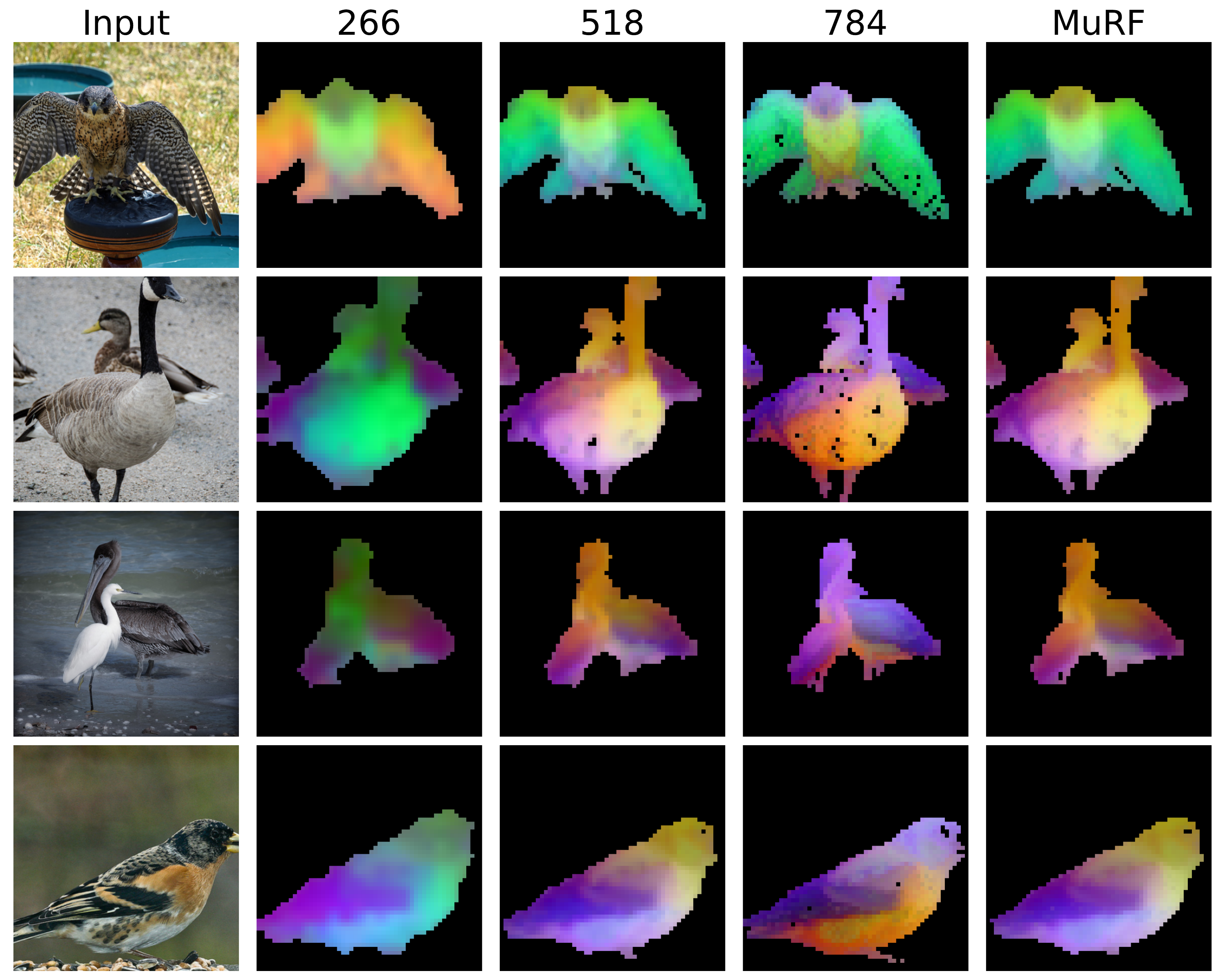}
    \includegraphics[width=0.48\textwidth]{images/pca_vis.pdf}
    \caption{Additional PCA visualization.}
    \label{fig:results_compare_supp}
\end{figure}

For all experiments, we used features from layer $7$, layer $9$ and layer $11$, and features from resolution $0.3$, $0.4$, $0.5$, $0.6$, $0.7$. The image is resized to the nearest multiple of patch size (14) before entering the model.

We use the IndexIVFFlat approximate nearest neighbor search algorithm from Faiss~\cite{johnson2019billion}. For all experiments, we used $n_{list}=512$ and $n_{probe} = 32$.

All experiments were conducted on a single server equipped with an NVIDIA A100-SXM4-80GB GPU and two AMD EPYC 7713 64-Core Processors. A full run of our method on a single MVTec AD v2 category takes less than 3 hours.

\section{Additional Visualization}

We provided additional visualization for Semantic Segmentation (Figure \ref{fig:seg_results_compare_supp}), Depth Estimation (Figure \ref{fig:depth_results_compare_supp}) and PCA Qualitative comparison (Figure \ref{fig:results_compare_supp}).

\appendix
\end{document}